\documentclass[letterpaper,12pt]{article}  % deGruyter

\usepackage{dgjournal}
\usepackage[authoryear,comma,longnamesfirst,sectionbib]{natbib} 

\newcommand*{\ARXIV}{}%  % when uncommented it uses the *.bbl file

\usepackage[title]{appendix}  %% simplified appendices
\usepackage{amsthm}    %% allows to use the \proof 

\newcommand{\CFilesBib}{Common.Files.Bib}

\usepackage{amsmath,amssymb,amscd,latexsym,dsfont,wasysym,bbm}
\usepackage{float}
\usepackage{color}
\usepackage{graphicx,subfigure}
\usepackage{multirow,multicol} 
\usepackage{verbatim}
\usepackage{psfrag}
\usepackage{comment}
\usepackage{makeidx}
\usepackage[]{algorithm}
\usepackage{algorithmicx}
\usepackage{algpseudocode}
\usepackage{cases}
 \usepackage{setspace}
\usepackage{pgfplots} %%%% added by LS for the errata 7.sept.2015

\newcommand{\mr}[1]{\mathrm{#1}}
\newcommand{\tnr}[1]{{\textnormal{#1}}}

\newcommand{\mc}[1]{\mathcal{#1}}
\newcommand{\mf}[1]{\mathsf{#1}}
 
\newcommand{\ms}[1]{\mathds{#1}}

\newcommand{\ov}[1]{\overline{#1}}

\newcommand{\bc}{\boldsymbol{c}}

\newcommand{\br}{\boldsymbol{r}}

\newcommand{\bp}{\boldsymbol{p}}

\newcommand{\bx}{\boldsymbol{x}}

\newcommand{\by}{\boldsymbol{y}}

\newcommand{\bzero}{\boldsymbol{0}}

\newcommand{\btheta}{\boldsymbol{\theta}}

\newcommand{\bxi}{\boldsymbol{\xi}}

\newcommand{\secref}[1]{Sec.~\ref{#1}}

\newcommand{\ie}{i.e.,~} 		%note the comma and the space (~)
\newcommand{\eg}{e.g.,~}	%note the comma and the space (~)
\newcommand{\argmin}{\mathop{\mr{argmin}}}

\newcommand{\set}[1]{\{#1\}}
\newcommand{\SET}[1]{\left\{#1\right\}}

\newcommand{\ld}{\ldots}

\newcommand{\e}{\mr{e}}

\newcommand{\PR}[1]{\Pr\SET{#1}}       	% Probability
\newcommand{\pdf}{f}            			% PDF. I vote for having the PDFs and the PMFs as italic because the pmf as \tr{P}_\bX(\bX) looks weird and also because I like to see the pmfs and the pdfs as functions, which we denote using italic letters (like the Q-function, f(t), s(t), h(t), etc.)

\newcommand{\IND}[1]{\ms{I}\big[{#1}\big]}   	% Indicator function
\newcommand{\Ex}{\ms{E}}     			% Expectation (AA).
\newcommand{\T}{^{\mf{T}}}            		% transpose
\newcommand{\dd}{\,\mr{d}}             		% LS: differentiation operator (\, added to make it nicer)
\newcommand{\mcN}{\mc{N}}

\newcommand{\mcY}{\mc{Y}}

\newcommand{\mfP}{\mf{P}}

\newcommand{\Real}{\mathbb{R}}		% R (The real,natural, etc. should used blackboard bold fonts... \mathbb{})
\newcommand{\matH}{\tnr{\textbf{H}}}
\newcommand{\matI}{\tnr{\textbf{I}}}

\usepackage[acronym,nonumberlist]{glossaries} % make a separate list of acronyms
\usepackage{hyperref}
\usepackage{empheq} % allows to put boxes around equations

\newacronym[\glsshortpluralkey=PDFs,\glslongpluralkey=probability density functions]{pdf}{PDF}{probability density function}
\newacronym[\glsshortpluralkey=CDFs,\glslongpluralkey=cumulative density functions]{cdf}{CDF}{cumulative density function}
\newacronym[\glsshortpluralkey=CCDFs,\glslongpluralkey=complementary cumulative density functions]{ccdf}{CDF}{complementary cumulative density function}
\newacronym[\glsshortpluralkey=PMFs,\glslongpluralkey=probability mass functions]{pmf}{PMF}{probability mass function}
\newacronym[]{lhs}{l.h.s.}{left-hand side}
\newacronym[]{rhs}{r.h.s.}{right-hand side} 

\newacronym[]{bicm}{BICM}{bit-interleaved coded modulation}
\newacronym[]{bicmid}{BICM-ID}{BICM with iterative demapping}
\newacronym[]{cm}{CM}{coded modulation}
\newacronym[]{tcm}{TCM}{trellis-coded modulation}
\newacronym[]{mlc}{MLC}{multi-level coding}
\newacronym[]{pam}{PAM}{pulse amplitude modulation}
\newacronym[]{bpsk}{BPSK}{binary phase shift keying}
\newacronym[]{qam}{QAM}{quadrature amplitude modulation}
\newacronym[]{16qam}{16-QAM}{16-points quadrature amplitude modulation}
\newacronym[]{psk}{PSK}{phase shift keying}
\newacronym[\glsshortpluralkey=LLRs,\glslongpluralkey=logarithmic likelihood ratios]{llr}{LLR}{logarithmic likelihood ratio}
\newacronym[]{oc}{OC}{operating characteristic}
\newacronym[]{dmp}{DMP}{discretized message passing}
\newacronym[]{mp}{MP}{message passing}
\newacronym[]{ep}{EP}{expectation propagation}
\newacronym[\glsshortpluralkey=MIs,\glslongpluralkey=mutual informations]{mi}{MI}{mutual information}
\newacronym[\glsshortpluralkey=GMIs,\glslongpluralkey=generalized mutual informations]{gmi}{GMI}{generalized mutual information}
\newacronym[]{eesm}{EESM}{exponential effective-SNR-mapping}
\newacronym[]{bicm-gmi}{BICM-GMI}{BICM generalized mutual information}
\newacronym[]{awgn}{AWGN}{additive white Gaussian noise}
\newacronym[]{bsc}{BSC}{binary symetric channel}
\newacronym[]{amc}{AMC}{adaptive modulation and coding}
\newacronym[]{csi}{CSI}{channel state information}
\newacronym[]{cqi}{CQI}{channel quality indicator}
\newacronym[]{kl}{KL}{Kullback-Leibler}
\newacronym[]{cmm}{CMM}{circular moment matching}
\newacronym[]{ga}{GA}{Gaussian approximation}

\newacronym[]{sp}{SP}{set-partitioning}
\newacronym[]{gsm}{GSM}{global system for mobile communications}
\newacronym[]{edge}{EDGE}{enhanced data rates for GSM evolution}
\newacronym[]{3gpp}{3GPP}{3rd generation partnership project}
\newacronym[]{umts}{UMTS}{Universal Mobile Telecommunication System}
\newacronym[]{lte}{LTE}{Long Term Evolution}
\newacronym[]{dvb}{DVB}{digital video broadcasting}
\newacronym[]{fdd}{FDD}{Frequency Division Duplexing}

\newacronym[\glsshortpluralkey=CCs,\glslongpluralkey=convolutional codes]{cc}{CC}{convolutional code}
\newacronym[\glsshortpluralkey=PCCCs,\glslongpluralkey=parallel concatenated convolutional codes]{pccc}{PCCC}{parallel concatenated convolutional code}
\newacronym[\glsshortpluralkey=TCs,\glslongpluralkey=turbo codes]{tc}{TC}{turbo code}
\newacronym{ldpc}{LDPC}{low-density parity-check}
\newacronym[]{ofdm}{OFDM}{orthogonal frequency-division multiplexing}
\newacronym[]{bep}{BEP}{bit-error probability}
\newacronym[]{wep}{WEP}{word-error probability}
\newacronym[]{sep}{SEP}{symbol-error probability}
\newacronym[]{pep}{PEP}{pairwise-error probability}
\newacronym[]{ttcm}{TTCM}{turbo-trellis coded modulation}
\newacronym[]{uep}{UEP}{unequal error protection}
\newacronym[\glsshortpluralkey=CENCs,\glslongpluralkey=convolutional encoders]{cenc}{CENC}{convolutional encoder}
\newacronym[]{mimo}{MIMO}{multiple-input multiple-output}
\newacronym[\glsshortpluralkey=SNRs,\glslongpluralkey=signal-to-noise ratios]{snr}{SNR}{signal-to-noise ratio}
\newacronym[\glsshortpluralkey=SINRs,\glslongpluralkey=the signal-to-interference-plus-noise ratios]{sinr}{SINR}{the signal-to-interference-plus-noise ratio}
\newacronym[]{msb}{MSB}{most-significative bit}
\newacronym[]{bcjr}{BCJR}{Bahl--Cocke--Jelinek--Raviv}
\newacronym[]{cbc}{CBC}{Colavolpe--Barbieri--Caire}
\newacronym[]{skr}{SKR}{Shayovitz--Kreimer--Raphaeli}
\newacronym[\glsshortpluralkey=SEDs,\glslongpluralkey=squared Euclidean distances]{sed}{SED}{squared Euclidean distance}
\newacronym[\glsshortpluralkey=EDs,\glslongpluralkey=Euclidean distances]{ed}{ED}{Euclidean distance}
\newacronym[\glsshortpluralkey=MEDs,\glslongpluralkey=minimum Euclidean distances]{med}{MED}{minimum Euclidean distance}
\newacronym[]{core}{CoRe}{constellation rearrangement}
\newacronym[]{pdl}{PDL}{parallel decoding of the individual levels}
\newacronym[\glsshortpluralkey=GCs,\glslongpluralkey=Gray codes]{gc}{GC}{Gray code}
\newacronym[]{brgc}{BRGC}{binary-reflected Gray code}
\newacronym[]{nbc}{NBC}{natural binary code}
\newacronym[]{fbc}{FBC}{folded-binary code}
\newacronym[]{bsgc}{BSGC}{binary semi-Gray code}
\newacronym[]{msp}{MSP}{modified set-partitioning}
\newacronym[]{ssp}{SSP}{semi set-partitioning}
\newacronym[]{fhd}{FHD}{free Hamming distance}
\newacronym[]{mfhd}{MFHD}{maximum free Hamming distance}
\newacronym[]{ods}{ODS}{optimal distance spectrum}
\newacronym[]{iud}{i.u.d.}{independent and uniformly distributed}
\newacronym[]{ud}{u.d.}{uniformly distributed}
\newacronym[]{iid}{i.i.d.}{independent, identically distributed}
\newacronym[]{ami}{AMI}{accumulated mutual information}
\newacronym[]{bico}{BICO}{binary-input continuous-output}
\newacronym[]{gh}{GH}{Gauss--Hermite}
\newacronym[]{gum}{GUM}{Gaussian--uniform mixture}
\newacronym[\glsshortpluralkey=BSs,\glslongpluralkey=base-stations]{bs}{BS}{base-station}
\newacronym[\glsshortpluralkey=MSs,\glslongpluralkey=mobile-stations]{ms}{MS}{mobile-stations}

\newacronym[]{phy}{PHY}{physical layer} 
\newacronym[]{rlc}{RLC}{Radio-Link control} 
\newacronym[]{ran}{RAN}{Radio Access Network} 
\newacronym[]{llc}{LLC}{logical link control} 
\newacronym[]{tcp}{TCP}{transmission control protocol} 
\newacronym[]{mac}{MAC}{media access control} 
\newacronym[]{fft}{FFT}{fast Fourier transform} 
\newacronym[]{ft}{FT}{Fourrier transform}
\newacronym[]{cf}{CF}{characteristic function} 
\newacronym[]{mgf}{MGF}{moment generating function} 
\newacronym[]{ee}{EE}{energy efficiency} 
\newacronym[]{eb}{EB}{energy per bit}
\newacronym[]{kkt}{KKT}{Karush--Kuhn--Tucker} 
\newacronym[]{mcs}{MCS}{modulation/coding scheme} 
\newacronym[]{fec}{FEC}{forward error correction}
\newacronym[]{arq}{ARQ}{automatic repeat request}
\newacronym[]{harq}{HARQ}{hybrid ARQ}
\newacronym[]{tarq}{TARQ}{truncated HARQ}
\newacronym[]{ir}{IR}{incremental redundancy}
\newacronym[]{rpr}{RR}{repetition redundancy}
\newacronym[]{rrharq}{RR-HARQ}{repetition redundancy HARQ}
\newacronym[]{irharq}{IR-HARQ}{incremental redundancy HARQ}
\newacronym[]{ack}{ACK}{positive acknowledgment}
\newacronym[]{nack}{NACK}{negative acknowledgment}
\newacronym[]{hol}{HoL}{head of the line}
\newacronym[]{crc}{CRC}{cyclic redundancy check}
\newacronym[]{dp}{DP}{dynamic programming}
\newacronym[]{gp}{GP}{geometric programming}
\newacronym[]{per}{PER}{packet error rate}
\newacronym[]{ber}{BER}{bit error rate}
\newacronym[]{op}{OP}{outage probability}
\newacronym[]{spa}{SPA}{saddle-point approximation}
\newacronym[]{mrc}{MRC}{maximum ratio combining}
\newacronym[]{mdp}{MDP}{Markov decision process}
\newacronym[]{lp}{LP}{linear programming}
\newacronym[]{pomdp}{POMDP}{partially observable Markov decision process}
\newacronym[]{psimdp}{PSI-MDP}{partial state information Markov decision process}
\newacronym[]{scpp}{SCPP}{stochastic shortest path problem}

\newacronym[]{forw}{frwd}{forward}
\newacronym[]{feed}{fdbk}{feedback}

\newacronym[]{mm}{MM-HARQ}{multi-message HARQ}
\newacronym[]{xp}{XP-HARQ}{cross-packet HARQ}
\newacronym[]{ts}{TS}{time-sharing}
\newacronym[]{sc}{SC}{superposition coding}
\newacronym[]{sbrq}{SBRQ}{systematic backward retransmission}
\newacronym[]{brq}{BRQ}{backward retransmission}
\newacronym[]{lharq}{L-HARQ}{layer-coded HARQ}
\newacronym[]{anlharq}{AoN-HARQ}{all-or-none L-HARQ}
\newacronym[]{vlharq}{VL-HARQ}{variable-length HARQ}

\newacronym[]{pp}{PPP}{point process}
\newacronym[]{ppp}{PPP}{Poisson point process}

\newacronym[]{fide}{FIDE}{F\'ed\'eration Internationale des \'Echecs}
\newacronym[]{fifa}{FIFA}{F\'ed\'eration Internationale de Football Association}
\newacronym[]{fivb}{FIVB}{F\'ed\'eration Internationale de Volleyball}
\newacronym[]{epl}{EPL}{English Premier League}
\newacronym[]{nhl}{NHL}{National Hockey League}
\newacronym[]{nfl}{NFL}{National Football League}

\newacronym[]{sg}{SG}{stochastic gradient}
\newacronym[]{lms}{LMS}{least mean squares}
\newacronym[]{rls}{RLS}{recursive least squares}
\newacronym[]{vss}{VSS}{variable step-size}
\newacronym[]{hfa}{HFA}{home-field advantage}
\newacronym[]{ha}{HA}{home advantage}
\newacronym[]{mov}{MOV}{margin of victory}
\newacronym[]{ac}{AC}{Adjacent Categories}
\newacronym[]{cl}{CL}{Cumulative Link}
\newacronym[]{rps}{RPS}{Ranked Probability Score}
\newacronym[]{mse}{MSE}{Mean Squared Error}
\newacronym[]{mmse}{MMSE}{Minimum Mean Squared Error}
\newacronym[]{rmse}{RMSE}{Root Mean Squared Error}
\newacronym[]{map}{MAP}{maximum a posteriori}
\newacronym[]{ml}{ML}{maximum likelihood}
\newacronym[]{loo}{LOO}{leave-one-out}
\newacronym[]{alo}{ALO}{approximate leave-one-out}
\newacronym[]{msd}{MSD}{mean-square-deviation}

\newacronym[]{svd}{SVD}{singular values decomposition}

\newacronym[]{skf}{SKF}{Simplified Kalman Filter}
\newacronym[]{vskf}{vSKF}{\emph{vector-covariance} Simplified Kalman Filter}
\newacronym[]{sskf}{sSKF}{\emph{scalar-covariance} Simplified Kalman Filter}
\newacronym[]{fskf}{fSKF}{\emph{fixed-variance} Simplified Kalman Filter}
\newacronym[]{kf}{KF}{Kalman Filter}
\newacronym[]{gelo}{G-Elo}{Generalized Elo}

\newacronym[]{tpb}{TPB}{tensor-product-basis}

\newtheorem{lemma}{Lemma}

\begin{document}

%\begin{frontmatter}

%------------------
%-- Paper title
\title{FIVB ranking: Misstep in the right direction
}

\author{
Salma Tenni\thanks{S.~Tenni was with INRS and McGill, Canada, e-mail: salma.tenni@mail.mcgill.ca.}, 
Daniel Gomes de Pinho Zanco\thanks{D.~G.~P.~Zanco is with Institut National de la Recherche Scientifique, Montreal, Canada, e-mail: Daniel.Zanco@inrs.ca.}, 
and Leszek Szczecinski\thanks{L. Szczecinski is with 
Institut National de la Recherche Scientifique, Montreal, Canada, e-mail: leszek.szczecinski@inrs.ca.}
}%

%-----------------
\ifdefined\ARXIV
\originalmaketitle  %% this shows the title and the authors' names when using the style: dgjournal.sty
\fi

\thispagestyle{empty}

\begin{abstract}
    This work presents and evaluates the ranking algorithm that has been used by \gls{fivb} since 2020. The prominent feature of the \gls{fivb} ranking is the use of the probabilistic model, which explicitly calculates the probabilities of the future matches results using the estimated teams' strengths. Such explicit modeling is new in the context of official sport rankings, especially for multi-level outcomes, and we study the optimality of its parameters using both analytical and numerical methods. We conclude that from the modeling perspective, the current thresholds fit well the data but adding the \acrfull{hfa} would be beneficial. Regarding the algorithm itself, we explain the rationale behind the approximations currently used and show a simple method to find new parameters (numerical score) which improve the performance. We also show that the weighting of the match results is counterproductive.
\end{abstract}

%%%%%%%%%%%%%%%%%%%%%%%%%%%%%%%%%%%%%%%%%%%%%%%%%%%%
\newpage
\section{Introduction}

The ranking of teams/players is one of the fundamental problems in competitive sports. It is used, for example, to declare the champion or to promote and relegate teams between leagues, and, in international competitions, to establish the composition of groups \eg in the qualification rounds of the FIFA World Cup. In other words, the ranking is a tool that allows the governing bodies to manage competitions by ``fairly'' evaluating the teams. In general terms, the ranking is meant to reflect the relative strength of the teams/players, and may be used for a quick assesment of the competitive landscape and, more consequentially, for tournament design where the strongest teams are scheduled to play in different group phase \citep{Csato24}. The formal evaluation of the ranking is based on its ability to predict the results of future matches \citep{Csato24}.

In this work, we present and evaluate the ranking algorithm, used by \acrfull{fivb}. Our study is motivated by the modern approach adopted by  \gls{fivb}, where six-level outcomes of the volleyball matches have explicit probabilistic models defined by \cite{fivb_rating}. To our knowledge, among main international sports,\footnote{Such as football, cricket, field hockey, tennis, volleybal, table tennis, golf, basketball, ice hockey, or tennis.} this is the first officially adopted ranking algorithm to use such an approach and, merely due to this fact, deserves attention in sport analytics. Of course, international volleyball is also a popular sport, and understanding its ranking strategy is interesting on its own merit.

%% what is special about the model
The \gls{fivb} ranking adopted in 2020 can be classified as \emph{power-ranking}, where teams are assigned a real-valued parameter called skills (also known as strength or power) and the teams are ranked (ordered) by sorting the skills. This approach departs from the more conventional ranking based on counting of the points associated with the results of the matches, and is often considered to be more ``fair''. 

The power-ranking approach was already adopted by \gls{fifa} to rank the Men's and Women's teams. The analysis of \cite{Szczecinski22a} revealed its weaknesses, where one of the criticisms was that the \gls{fifa} ranking, being based on the Elo ranking \citep{Elo08_Book}, inherits its main drawback, \ie the lack of an explicit probabilistic model of the match's outcomes \citep{Szczecinski20}.  From a statistical perspective, this lack of forecasting capability is, indeed, a significant drawback which makes it difficult to evaluate the ranking objectively. In this regard, the \gls{fivb} ranking proposes a radical and modern approach: each of the six outcomes of the volleyball match is assigned a probability that is calculated from the skills (known before the match) using the \gls{cl} model \citep[Ch.~9.1]{Tutz12_book}.

The objective of this work is to reveal the assumptions and simplifications used to derive the \gls{fivb} algorithm and to evaluate how well they are applied. In particular, we want to: i) propose the evaluation methodology of the \gls{fivb} ranking by casting it in a statistical framework, ii) show approximations used to derive the algorithm, and to iii) assess the optimality of the algorithm's parameters. In other words, we want to explain how and why the algorithm is working and whether it can be improved in the current general formulation.  In this regard, our work follows the approach of \citet{Szczecinski22a} which focused on evaluation and understanding of the currently used \gls{fifa} ranking. That is, we \emph{do not} want to propose/analyze new models or algorithms which would be rather in the spirit of many previous works in the area of sport ranking, \eg in \citep{Karlis08}, \citep{Egidi18}, \citep{Ntzoufras19}, \citep{Gabrio20}, \citep{Lasek20}, \citep{Szczecinski22}, or \citep{Macri24}.

Since there are many sports with multi-level outcomes (\ie taking more than two possible values), providing an understanding of how ranking algorithms may be constructed in such cases will be useful beyond the context of the \gls{fivb} ranking. We note that the previous works, \eg \citep{Egidi19} or \citep{Ntzoufras19}, already addressed the issue of modeling in volleyball. In our work, however, we analyze the ranking algorithm currently used by \gls{fivb} which, as we show, is not straightforwardly deduced from the model.

This work is organized as follows: in Sec.~\ref{Sec:Model}, we cast the ranking in the inference context, where the goal is to estimate the skills from the outcomes of the matches. This allows us to understand the approximate relationship between the probabilistic model and the \gls{fivb} ranking algorithm as such. The parameters of the model are assessed in Sec.~\ref{Sec:model.identification}, where, both the analytical and numerical approaches are applied to evaluate the importance of the model thresholds, the numerical scores used in the \gls{fivb} algorithms, the role of the \gls{hfa}, and the utility of weights associated with matches' categories. The parameters obtained in Sec.~\ref{Sec:model.identification}, are used in Sec.~\ref{Sec:Real.time.ranking} to assess the performance of the real-time ranking using the \gls{fivb} international matches.

We terminate the work in Sec.~\ref{Sec:Conclusions} summarizing the findings and showing the recommended changes to the \gls{fivb} ranking algorithm. Overall, we conclude that, from the statistical perspective, using an explicit probabilistic model to build a ranking algorithm is a step in the right direction. On the other hand, we qualify it as \emph{misstep} because many approximations lead to sub-optimality, which might have been easily avoided.

The repository \href{https://github.com/brbalab/FIVB}{https://github.com/brbalab/FIVB} contains code/data from which the results can be reproduced.

%%%%%%%%%%%%%%%%%%%%%%%%%%%%%%%%%%%%%%%%%%%%%%%%%%%%
\section{Model and ranking}\label{Sec:Model}
Consider the scenario, where, of the $M$ teams, two are selected to face each other in a match. The matches are indexed with $t=1,2,\ld,T$. Teams are indexed with a pair $(i_t, j_t)$, where $i_t, j_t \in \set{1, \ld, M}$. The team $i_t$ is called the home-team, and the team $j_t$ - the away-team. We keep this naming convention even when a match is played on a neutral venue.\footnote{A venue is neutral when, during an international tournament, the match is played in the country which is not the one of the team $i_t$ or $j_t$.}

The outcome of the $t$-th match $y_t\in\mcY$ is ordinal in nature, with meaning such as ``importance", which has no numerical value but may be ordered, and, for convenience, we index it with natural numbers $\mcY=\set{0,\ld, L-1}$ in decreasing order, \ie the outcome $y=0$ is the most important and $y=L-1$ is the least important one, where the importance is evaluated from the point of view of the home team $i_t$. Quite naturally, the most important outcome for the home team is the least important for the away team.

In particular, volleyball matches, which we will be interested in, produce $L=6$ possible outcomes ``3-0", ``3-1", ``3-2", ``2-3", ``1-3", ``0-3", where ``$k$-$l$" means that the home team won $k$ sets and the away team won $l$ sets; \ie the first three results mean that the home team won. For the purpose of the derivations, we use indices $y\in\mcY$, but it is easier to understand the meaning of the explicit results in the form ``$k$-$l$"; \ie the result $y_t=0$ corresponds to the outcome ``3-0", $y_t=1$ -- to ``3-1", and $y_t=5$ -- to ``0-3". We will use both and it should not lead to any confusion. 

%%%%%%%%%%%%%%%%%%%%%%%%%%%%%%
\subsection{Ranking as statistical inference}\label{Sec:ranking=inference}
The goal of the ranking is to order the teams and, to this end, we assume that they are characterized by intrinsic parameters called ``skills" $\theta_m\in\Real, m=1,\ld, M$. Then, by inferring the skills $\btheta=[\theta_1,\ld,\theta_M]\T$ from the observed outcomes of the matches $\by=[y_1,\ld,y_T]\T$, and ordering them, the ranking is naturally obtained.

Arguably the most popular probabilistic model of the relationship between the skills $\btheta$, and the random variable $Y_t$ (which models $y_t$) links the latter to the difference between the skills of the home and away teams:
\begin{align}
\label{Q.yz}
    \PR{Y_t=y| \btheta,\bx_t}
        &= \mfP_y(z_t),\\
    z_t &= \bx_t\T\btheta,
\end{align}
where $\mfP_y(z)$ is defined to strike balance between the complexity of the inference procedure and the modeling flexibility (that is, ability of the model \eqref{Q.yz} to fit the observations). Also, for compactness of notation, we introduce the scheduling vector
\begin{align}
    \bx_t=[x_{1,t},\ld,x_{M,t}]\T,    
\end{align}
where $x_{i_t,t}=1$, $x_{j_t,t}=-1$, and $x_{m,t}=0$ for $m\notin\set{i_t,j_t}$.

With the model defined, we may use conventional estimation strategies. For example, the \gls{map} estimate of $\btheta$ is obtained solving
\begin{align}
% \label{hat.theta.map}
    \hat\btheta 
%     & = \argmax_{\btheta} \log\pdf\Big(\btheta\big|\set{y_t,\bx_t}_{t=1}^T\Big)\\
    %& = \argmax_{\btheta} \log\prod_{t=1}^T\mfP_{y_t}(z_t)\pdf(\btheta),\\
\label{hat.theta.map.sum}
    & = \argmin_{\btheta} \sum_{t=1}^T\ell_{y_t}(z_t)+\rho(\btheta),
\end{align}
where, the negated log-score
\begin{align}
\label{log.score}
    \ell_{y}(z)&=-\log \mfP_y(z),
\end{align}
and the prior distribution of the skills is defined via $\rho(\btheta)=-\log\pdf(\btheta)$. The common assumption is to use the zero-mean Gaussian model for $\btheta$, and then
\begin{align}
\label{r(theta)=ridge}
    \rho(\btheta)  &= \frac{1}{2}\gamma\|\btheta\|^2 +\tnr{Const.}
\end{align}

A more general formulation is obtained by rewriting \eqref{hat.theta.map.sum} as
\begin{align}
\label{hat.theta}
    \hat{\btheta} 
    & = \argmin_{\btheta} J(\btheta)\\
    J(\btheta)&=\sum_{t=1}^T \ell^{\tnr{loss}}_{y_t}(\bx\T_t\btheta) + \rho(\btheta), %\frac{1}{2}\gamma \|\btheta\|^2,
\end{align}
where we use the ``loss function" $\ell^{\tnr{loss}}_y(z)$ which may, but does not need to, be the same as the log-score $\ell_y(z)$ in \eqref{log.score}. As we shall see, to simplify calculations, the loss function may be a proxy of the log-score. In this ``ordinal regression" formulation, $\rho(\btheta)$ is called a regularization function, and using \eqref{r(theta)=ridge} yields the well-known L2 regularization.

Note that, using $\gamma=0$ and $\ell^{\tnr{loss}}_y(z)=\ell_y(z)$ we obtain the conventional \gls{ml} estimation of the skills, which may be appropriate for ``sufficiently large'' $T$. However, with small/moderate $T$, it is prudent to use regularized form \eqref{hat.theta}.\footnote{Regularization is strictly necessary if there are teams whose matches finish with the extreme results ``3-0'' or ``0-3'' because then, without regularization, their skills tend to infinity.}

~

%%%%%%%%%%%%%%%%%%%%%%%%%%%%%%%%%%%%%%%%%%%%%
\noindent\textbf{Stochastic gradient ranking}

We provide \eqref{hat.theta} to lay out a conceptual reference framework in which we can analyze the models used for ranking. This regression approach can be applied with a moderate value of $T$ (when the skills do not change significantly). 

However, as also noted by \citet{Csato24}, practical considerations of simplicity and transparency are more important in the context of sport ranking than the exact formulation. Thus, the online (real-time) ranking, where the skills are updated after each match, is more common in practice and is obtained by solving \eqref{hat.theta} using the \gls{sg} algorithm, defined as follows:
\begin{align}
    \hat\btheta_{t+1}
    &=
    \hat\btheta_{t} - \mu \big[\nabla_{\btheta} \ell^{\tnr{loss}}_{y_t}(\bx_t\T\btheta)\big]_{\btheta=\hat{\btheta}_t}\\
\label{SG.btheta}
    &=
    \hat\btheta_{t} - \mu \bx_t \dot\ell^{\tnr{loss}}_{y_t}(\bx_t\T\hat\btheta_{t}),
\end{align}
where $\mu$ is the adaptation step, and
\begin{align}
\label{dot.ell.loss}
    \dot\ell^{\tnr{loss}}_{y}(z) & = \frac{\dd}{\dd z}\ell^{\tnr{loss}}_{y}(z).
\end{align}
The algorithm is initialized with $\btheta_0$, \eg with $\btheta_0=[0,\ld,0]\T$.

When $\ell^{\tnr{loss}}_{y}(z)$ is convex in $z$, for sufficiently large $t$ and appropriately chosen $\mu$, the solution of \eqref{SG.btheta} approximates ``well" $\hat\btheta$. The adaptation step, $\mu$, trades off the convergence speed (which tells how quickly, with $t$, $\hat\btheta_t$ approaches $\hat\btheta$) against the accuracy (which measures how far $\hat\btheta_t$ is from $\hat\btheta$). Although this is, admittedly, a vague statement, the precise analysis of the \gls{sg} solution is not trivial even if some light is shed on this issue, \eg in \citep{Aldous17}, \citep{Jabin20}, \citep{Szczecinski22a}, \citep{Zanco23}. In fact, \eqref{SG.btheta} may be seen as an approximation of a nonlinear Kalman filter, which estimates skills at time $t$ from previous observations $y_1,\ld, y_{t-1}$ \citep[Sec.~3.3]{Szczecinski22a}. %In this perspective, the \gls{sg} algorithm tracks the skills $\btheta_t$, as they change through time.

In this perspective, the skills $\hat\btheta_t$ are approximate solutions to the \gls{ml} estimation of the skills $\btheta$: the approximation is due to the use of the stochastic gradient and due to the use of the loss function, which may approximate the log-score.
\\

%%%%%%%%%%%%%%%%%%%%%%%%%%%%%%%%%%%%%%%%%%%%%
\noindent\textbf{Scale}

It turns out that the skills estimates, $\hat{\theta}_m$ ($\hat\theta_{m,t}$ in case of the \gls{sg} algorithm \eqref{SG.btheta}), may be quite small and therefore, to place them in a comfortable range (\eg for visual interpretation by the users), we may multiply them by an arbitrarily chosen scale $s>0$, \ie making the change of variables $\btheta'=s\btheta$. This scale change can be made directly on the final solutions in \eqref{hat.theta}, as $\hat\btheta'=s\hat\btheta$ or in \eqref{SG.btheta}, as $\hat\btheta'_t=s\hat\btheta_t$. In fact, the latter can be integrated into the recursive equation yielding
\begin{align}
\label{SG.btheta.scaled}
    \hat\btheta'_{t+1}
    &=
    \hat\btheta'_{t} - \mu s \bx_t \dot\ell^{\tnr{loss}}_{y_t}(\bx_t\T\hat\btheta'_{t}/s).
\end{align}

%%%%%%%%%%%%%%%%%%%%%%%%%%%%%%%%%%%%%%%%%%%%%
\subsection{Integrating exogenous variables}\label{Sec:Exagenous.variables}

We assumed that the probabilistic model \eqref{Q.yz} depends only on $z_t=\bx\T_t\btheta$ and the outcome $y_t$. However, a more general approach may be used in which other exogenous variables affect the model.\\ 

%%%%%%%%%%%%%%%%%%%%%%%%%%%%%%%%%%%%%%%%%%%%%
\noindent\textbf{\Acrfull{hfa}}

For example, we may want to take into account the \gls{hfa} using a binary variable $h_t\in\set{0,1}$ that indicates whether the match $t$ is played on the home venue, \ie in the country of the team $i_t$ (then $h_t=1$), or is played in the neutral venue (then $h_t=0$). The popular model relies on boosting the skills of the home team, or, equivalently, on increasing the values of $z_t$ by the \gls{hfa} parameter $\eta$, \ie
\begin{align}
\label{hfa.loss.definition}
    \ell^{\tnr{loss}}_{y_t,h_t}(z_t)= \ell^{\tnr{loss}}_{y_t}(z_t + h_t \eta);
\end{align}
the exogenous variable $h_t$ is shown as a subscript, and $\eta$ is a part of the model. \\

%%%%%%%%%%%%%%%%%%%%%%
\noindent\textbf{Weighting/matches importance}

Using the same notation, we may modify the loss function via heuristics, such as weighting 
\begin{align}
\label{weighted.loss.definition}
    \ell^{\tnr{loss}}_{y_t,v_t}(z_t) &= \xi_{v_t}\ell^{\tnr{loss}}_{y_t}(z_t)
\end{align}
where $v_t$ is a categorical variable associated with  match $t$, $v_t\in\set{0,\ld,K-1}$, and the weight $\xi_{v_t}\ge 0$ allows us to modulate the relative importance of the term $\ell^{\tnr{loss}}_{y}(z)$: the smaller $\xi_{v_t}$ is, the less impact the pair $(z_t,y_t)$ will have on the solution $\hat\btheta$.  

In the ranking context, $v_t$ is called the ``prestige" of the match (in the \gls{fivb} ranking) or its ``importance" (in the \gls{fifa} ranking). 

The weights $\bxi=[\xi_0,\ld,\xi_{K-1}]$ are then subjectively defined by experts.
Note that, multiplying all the terms under optimization \eqref{hat.theta} by a positive constant does not change the optimization results, therefore, without loss of generality, we may set $\xi_0=1$.\footnote{This amounts to dividing all terms by $\xi_0$, which would also affect the regularization function $\rho(\btheta)$. However, this is inconsequential and amounts to using a regularization coefficient $\gamma/\xi_0$, see \eqref{r(theta)=ridge}.} Although subjective weighting is not uncommon in statistical literature, see \eg \citep{Hu01}, \citep{Ley19}, it is possible to evaluate the weighting objectively. For example, \cite{Szczecinski22a} concluded that the use of weighting is counterproductive in the \gls{fifa} ranking. 

%%%%%%%%%%%%%%%%%%%%%%%%%%%%%%%%%%%%%%%%%%%%%%%%%%%%%%%%%
\subsection{Current FIVB ranking algorithm}\label{Sec:FIVB.definition}

The \gls{fivb} ranking defines \eqref{Q.yz}
\begin{align}
    \mfP^{\tnr{FIVB}}_y(z)&=\mfP_y(z;\bc^{\tnr{FIVB}})\\
\label{Qy(z).FIVB}
    \mfP_y(z;\bc)
    &= \Phi(z+c_{y}) - \Phi(z+c_{y-1}),\quad y=0,\ld, L-1,
\end{align}
where $\Phi(z)=\frac{1}{\sqrt{2\pi}}\int_{-\infty}^z\exp(-0.5v^2)\dd v$ is the \gls{cdf} of a zero-mean, unit-variance Gaussian distribution. 

The model \eqref{Qy(z).FIVB} is an extension of the Bradley-Terry model for binary outcomes \citep{Bradley52} and was also used with ternary outcomes by \citet{Glen60}. The generalization to multi-level outcomes in \eqref{Qy(z).FIVB} is known as the ordinal probit \cite[Ch.~8.3]{Agresti13_book}; we call it a ``\gls{cl} model'' \citep[Ch.~9.1]{Tutz12_book} which is more general allowing us to use non-Gaussian \gls{cdf}s in the model.

The model is parameterized with \emph{thresholds} $\bc=[c_{-1}, c_{0},\ld, c_{L-1}]$, which are monotonically increasing with $l$, and, to simplify the discussion, may also be assumed symmetric, \ie 
\begin{align}\label{c.l.are.symmetric}
    c_l = -c_{L-2-l},
\end{align}
and we always set the first, and the last thresholds as $c_{-1}=-c_{L-1}=-\infty$. Also, for even $L$, we have $c_{\frac{L}{2}-1}\equiv 0$. In particular, in the \gls{fivb} model, with $L=6$, we always have $c_2\equiv 0$, and $-c_{-1}=c_{5}=\infty$ and thus only two parameters can be set independently: $c_0$ and $c_1$.

The \gls{fivb} ranking defines the thresholds as follows:
\begin{align}
\label{bc.FIVB}
    &c^{\tnr{FIVB}}_{0}=-c^{\tnr{FIVB}}_{4}=-1.06,& &c^{\tnr{FIVB}}_{1}=-c^{\tnr{FIVB}}_{3}=-0.394,&
    &c^{\tnr{FIVB}}_{2}=0.&
    % \bc=[-\infty, -1.06, -0.394, 0.0 , 0.394, 1.06, \infty].
\end{align}

\begin{figure}[tb]
    \centering
    \includegraphics[width=1.0\linewidth]{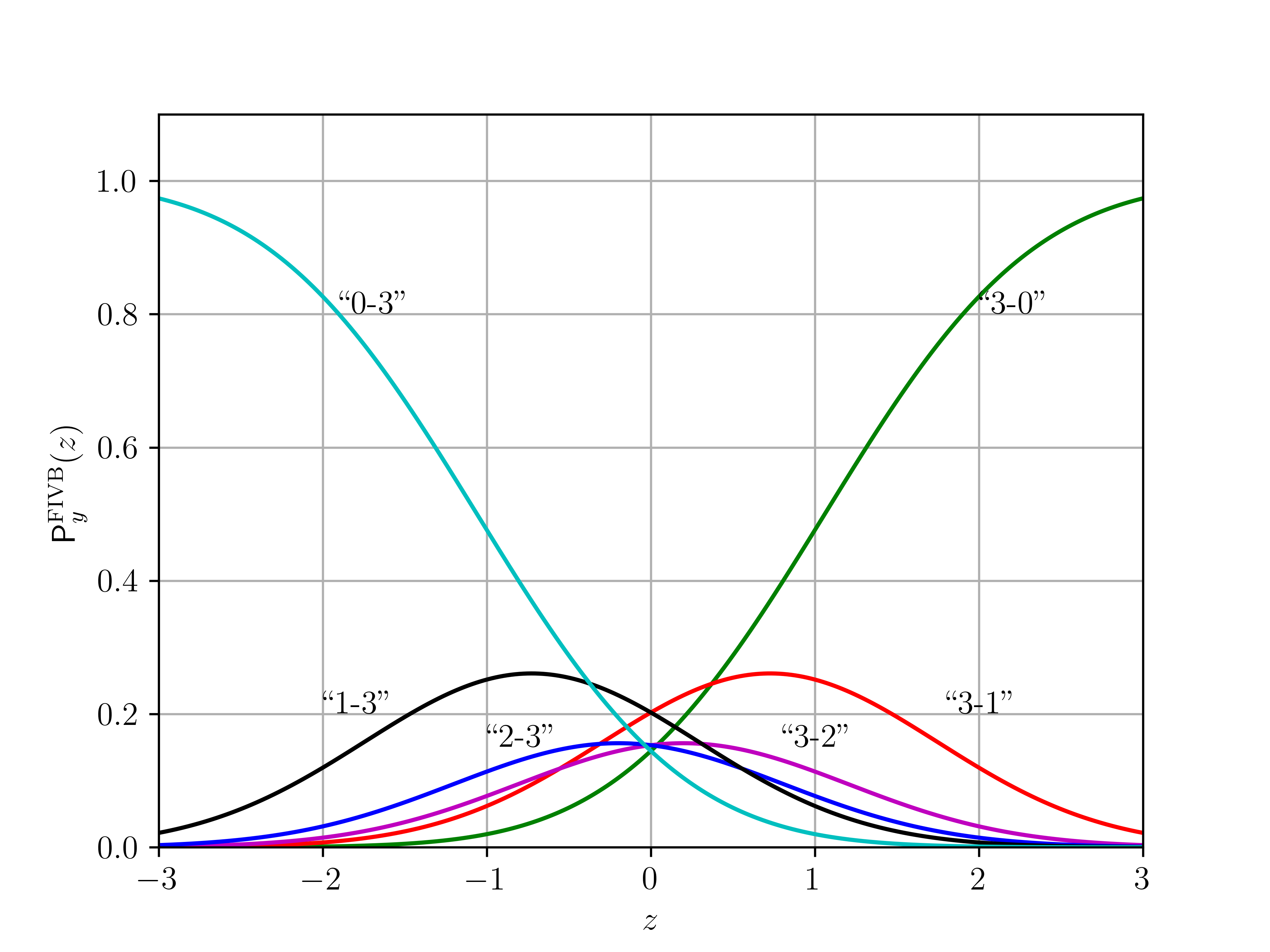}
    \caption{Functions $\mfP^{\tnr{FIVB}}_y(z)$ in the \gls{fivb} ranking algorithm, where $y=0$ corresponds to the result ``3-0", $y=1$ to ``3-1", etc.}
    \label{fig:Qy(z)}
\end{figure}

The symmetry of the thresholds $c^{\tnr{FIVB}}_y=-c^{\tnr{FIVB}}_{L-2-y}$, and of the \gls{cdf}, $\Phi(z)=1-\Phi(-z)$, yields the symmetric forms $\mfP^{\tnr{FIVB}}_{y}(z)=\mfP^{\tnr{FIVB}}_{L-1-y}(-z)$, as can be appreciated in Fig.~\ref{fig:Qy(z)}.
One of the properties of the \gls{cl} model is that the probability of the outcome ``not less important than $y$'' is calculated as
\begin{align}
\label{Pr.Y<=y}
    \PR{Y\le y|z} = \sum_{l=0}^{y}\mfP^{\tnr{FIVB}}_l(z)=\Phi(z+c^{\tnr{FIVB}}_y).
\end{align}

%%%%%%%%%%%%%%%%%%%%%%%%%%%%%%%%%%%%%%%%%%%%%%%%%%%
%%%%%%%%%%%%%%%%%%%%%%%%%%%%%%%%%%%%%%%%%%%%%%%%%%%

With the model defined, the \gls{fivb} ranking algorithm \citep{fivb_rating} estimates the skills as follows:
\begin{align}
\label{FIVB.algorithm}
    \btheta_{t+1} & 
    =
    \btheta_t - \mu s \xi_{v_t}\bx_t g^{\tnr{FIVB}}_{y_t}(z_t/s),\\
    \label{G.FIVB}
    g^\tnr{FIVB}_{y}(z) 
    &= \check{r}(z)- r^\tnr{FIVB}_y,
\end{align}
where the adaptation step is $\mu=0.01$, the scale is $s=125$, the weights $\xi_v$ are defined in Table~\ref{tab:FIVB.xi}, and
$r^\tnr{FIVB}_y$ is the \emph{numerical} score assigned to the outcome $y$, see Table~\ref{tab:numerical.values.FIVB}. It may be surprising because, as we emphasized previously, ordinal outcomes have no numerical value. This is still true: in fact, as we will see in Sec.~\ref{Sec:Implicit.Loss.Function}, variables $r^\tnr{FIVB}_y$ are merely auxiliary parameters defining the loss function $\ell^{\tnr{loss}}_y(z)$, which is not the same as the log-score \eqref{log.score}. 

The probabilistic model \eqref{Qy(z).FIVB} is used to calculate the expected value of $r^\tnr{FIVB}_{Y}$ (for a given $z$):
\begin{align}
\label{check.vt}
    \check{r}(z) &= \Ex_{Y|z}[r^\tnr{FIVB}_Y]=\sum_{y\in\mcY} r^\tnr{FIVB}_y \mfP^{\tnr{FIVB}}_y(z)\\
\label{check.vt.reorganized}
    &= \sum_{y=0}^{L-2}(r^\tnr{FIVB}_y-r^\tnr{FIVB}_{y+1})\Phi(z+c^{\tnr{FIVB}}_y) + r^\tnr{FIVB}_{L-1}.
\end{align}

%%%%%%%%%%%%%%%%%%%%%%%%%%%%%%%%%%%%%%%%%%%%%%%%%%%
\begin{table}[h]
    \centering
    \begin{tabular}{c|c|c}
       $v$ &  $\xi_v$   & Description\\
       \hline
    $0$ & $1.00$ & Official events of Continental Confederations\\
    \hline
    $1$ & $1.75$ & Confederations' Championship qualifying\\
    \hline
    $2$ & $2.00$ & FIVB Challenger Cup\\
    \hline
    \multirow{2}{*}{$3$} & \multirow{2}{*}{$3.50$} &  Olympic matches qualifying, \\
      %&  & FIVB World Cup (not played since 2019)\\
      &  & Confederations' Championship\\
      \hline
    $4$ & $4.00$ & FIVB Nations League\\
    \hline
    $5$ & $4.50$ & FIVB World Championship\\
    \hline
    $6$ & $5.00$ & Olympic matches    
    \end{tabular}
    \caption{The weighting coefficients adopted from the \gls{fivb} ranking. Note that we normalize $\xi_v$ by using $\xi_v=1.0$.}
    \label{tab:FIVB.xi}
\end{table}

%%%%%%%%%%%%%%%%%%%%%%%%%%%%%%%%%%%%%%%%%%%%%%%%%%%
\begin{table}[h]
    \centering
    \begin{tabular}{c|c|c|c|c|c|c}
         $y$   &  ``3-0"   &   ``3-1"  &   ``3-2"  &  ``2-3" & ``1-3"   & ``0-3"\\
         \hline
         $r^\tnr{FIVB}_y$ &  2.0  & 1.5  &  1.0 & -1.0 & -1.5 & -2.0\\
    \end{tabular}
    \caption{Numerical scores $r^\tnr{FIVB}_y$ assigned to the outcomes $y$ in the \gls{fivb} ranking algorithm.}
    \label{tab:numerical.values.FIVB}
\end{table}
%%%%%%%%%%%%%%%%%%%%%%%%%%%%%%%%%%%%%%%%%%%%%%%%%%%

The \gls{fivb} algorithm defined by \eqref{FIVB.algorithm}-\eqref{check.vt} is meant to be compatible with the formulation in \eqref{SG.btheta.scaled} but, of course, the official \gls{fivb} presentation  \citep{fivb_rating} uses a different notation. For example, instead of defining the step, the scale and the weights, the \gls{fivb} ranking defines their product $\mu s \xi_v$, which is required in \eqref{FIVB.algorithm}; we explain this in detail in Appendix~\ref{Sec:FIVB.notation}. %In fact, the \gls{fivb} description specifies directly the value of $8\mu s \xi_v$ (which divided by $8$ yields $\mu s \xi_v$).

The \gls{fivb} ranking has an additional rule, where, at the end of the year, each team $m$ that did not participate in any competition has the value of its skills reduced $\theta_{m,t}\leftarrow \theta_{m,t} -50$. The objective is to discourage match avoidance to preserve the value of the skills (and thus the ranking position). But these are heuristics that cannot be easily put in the statistical framework, and thus we do not model them. 

%%%%%%%%%%%%%%%%%%%%%%%%%%%%%%%%%%%%%%%%%%%%%%%%%%%
\subsection{Data}\label{Sec:data}
To evaluate the \gls{fivb} ranking algorithm, we use the matches played by the men's national teams and published on the \gls{fivb} website \citep{fivb_rating}. We only consider the matches since January 1, 2021, as, before that date, due to the Covid-19 pandemic, many matches were eliminated and some had the weighting $\xi_v$ incompatible with the official description.\footnote{For example, the matches played in January 2020 used the weight $\xi_v=2.50$ which is not specified in the ranking.} The results were collected till the end of December 2023 (with no matches after October 2023).

Since the \gls{fivb} website publishes the result $y_t$ of the match $t$ and the change in the value of skills, $\theta_{m,t+1}-\theta_{m,t}$, we can find the value of $\xi_{v_t}$ and infer the match category $v_t$ from Table~\ref{tab:FIVB.xi}. To establish the venue of the match (which affects $h_t$), we relied on Wikipedia pages describing the international volleyball events that we paired with matches from the \gls{fivb} website.\footnote{These results were non exhaustively validated by comparing with the venues of main international events shown on the \gls{fivb} website (such as Nations League, Challenger Cup or World Championship).}

Moreover, we remove
\begin{itemize}
    \item The matches in which, the \gls{fivb} ranking displays increments with small absolute value $|\theta_{m,t+1}-\theta_{m,t}| \in \set{ 0, 0.01}$.\footnote{There were $67$ matches with the absolute value of the increment equal to $0.01$, and $33$ with the increment equal to $0.0$. For example, the Oct. 8, 2023 match China-Poland had the increment equal to $0.01$, and the 24 Sept., 2023, USA-Canada match had the increment equal to $0.0$.} This may happen, \eg when players could not obtain visas, and sometimes it is explained on the \gls{fivb} website (\eg in the case of the Iranian team playing in the USA in 2023) but, in other cases, the non-standard values of $\xi$ are left unexplained. Since these matches do not contribute to the change in the ranking, excluding them from considerations is consistent with the spirit of the \gls{fivb} ranking.
    \item The forfeited matches we could identify,\footnote{Including: i) Denmark's matches in Jan. 2021, ii) Uzbekistan's and Pakistan's matches in July 2023, and iii) Mongolia's matches in Aug. 2023.} Even if \gls{fivb} treats a forfeited match as a ``$3-0$" win, in our opinion, the match which did not take place does not reflect on the strength of the team, and thus we do not use it as information for ranking.
\end{itemize}

Then,  we have a total of $M=102$ teams and $T=1151$ matches. To characterize the results, we count matches with result $y_t=y$, played on the  neutral- and home-venues
\begin{align}
\label{k.ntr.l}
    k^{\tnr{ntr}}_y&=\frac{1}{2}\sum_{t=1}^T\big(\IND{y_t=y, h_t=0} + \IND{y_t=L-1-y, h_t=0}\big),\\ 
    k^{\tnr{hfa}}_y&=\sum_{t=1}^T\IND{y_t=y, h_t=1}
\end{align}
and show them in Table~\ref{tab:empirical.frequencies}.

Note that in neutral-venue matches, there is no distinction between the home/away teams, so the number of results ``$k-l$" (denoted as $y$) and ``$l-k$" (denoted as $L-1-y$) should be equal: this is why we count all these results in \eqref{k.ntr.l}. Although this produces a fractional value $k^{\tnr{ntr}}_y$, this formalism simplifies the notation, and we guarantee that $k^{\tnr{ntr}}_y=k^{\tnr{ntr}}_{L-1-y}$.
%%%%%%%%%%%%%%%%%%%%%%%%%%%%%%%%%%%%%%%%%%%%%%%%%%%
\begin{table}[h]
    \centering
    \begin{tabular}{c|c|c|c|c|c|c||c}
         $y$   &  ``3-0"   &   ``3-2"  &   ``3-1"  &  ``1-3" & ``2-3"   & ``0-3" & Total\\
         \hline
         $k^{\tnr{ntr}}_y$&  203.5  &     117.5  &  59.5 &     59.5  &   117.5 &  203.5 & 761\\
                  \hline
         $k^{\tnr{hfa}}_y$&  135  &     64  &  29 &     33  &   45 &  84 & 390\\
    \end{tabular}\\
    \caption{Numbers of the \gls{fivb} matches with outcomes $y\in\mcY$, played on the neutral- and home- venues, and their totals.}
    \label{tab:empirical.frequencies}
\end{table}
%%%%%%%%%%%%%%%%%%%%%%%%%%%%%%%%%%%%%%%%%%%%%%%%%%%

%%%%%%%%%%%%%%%%%%%%%%%%%%%%%%%%%%%%%%%%%%%%%%%%%%%
\subsection{Implicit loss function}\label{Sec:Implicit.Loss.Function}

Knowing the probabilistic model, we can derive the \gls{sg} algorithm setting $\ell_y^{\tnr{loss}}(z)=\ell_y(z)$, \ie using the log-score as a loss function with
\begin{align}
\label{log.score.FIVB}
    \ell_y(z) &= - \log\big(\Phi(z+c_y)-\Phi(z+c_{y-1})\big),
\end{align}
and finding the derivative of the latter
\begin{align}
\label{dot.log.score.FIVB}
    \dot\ell_y(z) &= - \frac{\mcN(z+c_y)-\mcN(z+c_{y-1})}{\Phi(z+c_y)-\Phi(z+c_{y-1})},
\end{align}
where we use the Gaussian \gls{pdf}
\begin{align}
    \mcN(z) = \dot\Phi(z)=\frac{1}{\sqrt{2\pi}}\exp(-0.5z^2).
\end{align}

Quite obviously, plugging $\dot\ell_y(z)$ shown in \eqref{dot.log.score.FIVB}, into the \gls{sg} algorithm \eqref{SG.btheta.scaled}, \emph{does not} produce the \gls{fivb} ranking \eqref{FIVB.algorithm}. Thus, the \gls{fivb} ranking does not solve the \gls{ml} or \gls{map} problem. 

To understand what problem it \emph{does} solve, we note that, since the \gls{fivb} ranking \eqref{FIVB.algorithm} has the structure of the \gls{sg} optimization \eqref{SG.btheta.scaled}, we may treat the function $g^{\tnr{FIVB}}_y(z)$ used by \eqref{FIVB.algorithm} as a derivative of an implicit (that is, not explicitly defined) loss function $\ell^{\tnr{FIVB}}_y(z)$, \ie $g^{\tnr{FIVB}}_y(z)=\dot\ell^{\tnr{FIVB}}_y(z)$, and, the latter can be unveiled through the integration of $g^{\tnr{FIVB}}_y(z)$, \ie
\begin{align}
    \ell^{\tnr{FIVB}}_y(z)
    &=\int_{-\infty}^z g^{\tnr{FIVB}}_y(u) \dd u
    = \int_{-\infty}^z \big[\check{r}(u) - r^\tnr{FIVB}_y \big] \dd u,\\
\label{implicit.loss.FIVB}
    &= 
    \sum_{l=0}^{L-2}(r^{\tnr{FIVB}}_l - r^{\tnr{FIVB}}_{l+1})\psi(z+c^{\tnr{FIVB}}_l) 
    + (r^\tnr{FIVB}_{L-1}-r^\tnr{FIVB}_y) z + \tnr{Const.}
    % \psi(z+c_{0}) r_0 \\
    % &+
    % \sum_{l=1}^{L-2} \big[ \psi(z+c_{l})-\psi(z+c_{l-1}) \big] r_l\\
    % &+
    % \big[ z - \psi(z+c_{L-2}) \big] r_{L-1}
    % - r^\tnr{FIVB}_y z 
    % +\tnr{Const.}\\
\end{align}
where \eqref{implicit.loss.FIVB} is obtained from \eqref{check.vt.reorganized} and
\begin{align}
    \psi(z) &= \int_{-\infty}^z \Phi(u)\dd u
    =
    \Phi(z)z + \mcN(z).
\end{align}

To understand why the \gls{fivb} ranking algorithm does not use directly the log-score \eqref{log.score.FIVB}, two issues with \eqref{dot.log.score.FIVB} should be considered. 

The first is of numerical nature, because, for large $|z|$, both the numerator and the denominator in \eqref{dot.log.score.FIVB} tend to zero which requires a careful implementation.
%\footnote{Although large values $|z|$ are unlikely to appear in real-world ranking implementation, they do appear when solving optimization \eqref{hat.theta}. In any case, it is preferable to use a numerically stable formulation.
%
%We note that even the implementation of \eqref{log.score.FIVB} is difficult because, in floating-point implementation, for large $z>0$, $\Phi(z+c_y)$ and $\Phi(z+c_{y-1})$ tend to $1$ and the precision of their difference is constrained by the number of bits in the mantissa representation. We may then exploit the symmetry of the \gls{cdf} to calculate it only for negative arguments, \ie $\Phi(z+c_y)-\Phi(z+c_{y-1})=\Phi(-z-c_{y-1})-\Phi(-z-c_y)$. 
%
%However, even with this trick, with growing $|z|$, we need to use the log-domain implementation
% \begin{align}
%     \log\big(\Phi(z+c_y)-\Phi(z+c_{y-1})\big)
%     &=
%     \Psi(z+c_y)+\log\Big[1-\e^{\Psi(z+c_{y-1})-\Psi(z+c_y)}\Big]
% \end{align}
% where $\Psi(z)=\log\Phi(z)$, available in numerical software packages, is calculated without explicit evaluation of the \gls{cdf}.

% This is merely done to make the calculation feasible. However, the resulting formulas are too complex to produce a practical ranking algorithm. 
% }

The second problem is that \eqref{dot.log.score.FIVB} has a relatively complicated form without obvious interpretation. In that regard, \eqref{G.FIVB} has a practical advantage: the skills $\btheta_t$ are updated using the difference between the observed and expected numerical scores. Note that this is also the usual interpretation of the well-known Elo algorithm \citep{Elo08_Book}. 

We therefore conjecture that the \gls{fivb} ranking algorithm was designed to be simple and understandable, which is a typical requirement in the sport ranking \citep[Sec.~1]{Csato24}. The potential drawback is the sub-optimality of the ranking results, which we will evaluate in this work.\\

%%%%%%%%%%%%%%%%%%%%%%%%%%%%%%%%%%%%%%%%%%%%%%%%%%%
\noindent
\textbf{Convergence}

A minor point is to obtain a guarantee that, for a sufficiently small $\mu$, the \gls{fivb} algorithm \eqref{FIVB.algorithm} converges, for which we need the following:
\begin{lemma}\label{lemma:1}
    The implicit loss function, $\ell^{\tnr{FIVB}}_y(z)$ is convex in $z$.\\
\textbf{Proof}:\\
The second derivative of implicit loss $\ell^{\tnr{FIVB}}_y(z)$ is positive if the first derivative $g^{\tnr{FIVB}}_y(z)=\check{r}(z)-r^\tnr{FIVB}_y$ increases monotonically in $z$. The latter holds because \eqref{check.vt.reorganized} is monotonically increasing if $r^\tnr{FIVB}_{y+1}<r^\tnr{FIVB}_y$, which is true from Table~\ref{tab:numerical.values.FIVB}.
\end{lemma}

We note that condition $r^\tnr{FIVB}_{y+1}<r^\tnr{FIVB}_y$ is sufficient but not necessary, \ie it can be violated, and yet we can still obtain a convex function $\ell^{\tnr{FIVB}}_y(z)$ (see examples in Sec.~\ref{Sec:Numerical.optimization.of.r}). The fact that this is possible only reinforces the idea that numerical scores are auxiliary parameters and should not be thought of as a \emph{value} of the match outcome (because the latter simply do not exist). % (although this would lead to somewhat counterintuitive results, with the numerical score $r^\tnr{FIVB}_y$ not decreasing with $y$)

%%%%%%%%%%%%%%%%%%%%%%%%%%%%%%%%%%%%%%%%%%%%%%%%%%%
\section{Model identification}\label{Sec:model.identification}

% For the model identification, we assume that the form of the functions is predefined and our goal is to question the rationale behind the choice of the parameters. In particular, we will a) find the thresholds $\bc$ and the \gls{hfa} parameter $\eta$, b) find numerical scores $\br$ which, for a given $\bc$, produce $\ell^{\tnr{FIVB}}_y(z)$ which approximates ``well"  the log-score $\ell_y(z)$, and c) assess the importance of using the weights $\bxi$.

In this part of the work,  we assume that 
\begin{itemize}
    \item The probabilistic model  \eqref{Qy(z).FIVB} is predefined, and we need to define the thresholds $\bc$ and, eventually, the \gls{hfa} coefficient $\eta$.
    \item The loss function is defined via $\ell^{\tnr{FIVB}}_y(z)$ in \eqref{implicit.loss.FIVB}, and we need to define the suitable numerical scores $\br=[r_0,\ld, r_{L-1}]$ which are attributed to the outcomes, and
    \item The weights $\bxi$ are used in the algorithm via \eqref{weighted.loss.definition}, and we want to assess their usefulness.
\end{itemize}

In order to assess and/or optimize the parameters of the model, we need a well-defined criterion.

%%%%%%%%%%%%
\subsection{Model identification via cross-validation}\label{Sec:Cross-validation}

Inference \eqref{hat.theta} may be done if we define (the form of) the loss and regularization functions, as well as, if we find their parameters $\bp$ (also called hyper-parameters, in the machine learning language) that affect all the functions describing the models, \ie we can write $\ell^{\tnr{loss}}(z)\equiv \ell^{\tnr{loss}}_y(z;\bp)$, $\ell_y(z)\equiv\ell_y(z;\bp)$ and $\rho(\btheta)\equiv \rho(\btheta;\bp)$. 

A well-known approach to finding the hyper-parameters, particularly suitable for relatively small data sets, relies on \gls{loo} cross-validation \citep[Ch.~2.9]{Hastie_book}, where we remove one observation $y_t$, and we verify how well the results \eqref{hat.theta}, denoted as $\hat\btheta_{\backslash{t}}$, match the removed data using a metric $\ell^{\tnr{val}}_{y_t}(\bx_t\T\hat\btheta_{\backslash{t}})$; for the latter, most often we use the log-score $\ell^{\tnr{val}}_{y}(z)=\ell_y(z)$. This is repeated for all $T$ samples and may be defined as follows \citep[Ch.~2.9]{Hastie_book} 
\begin{align}
\label{optimize.hyper}
    \hat\bp &= \argmin_{\bp} U(\bp)\\
\label{U(p).define}
    U(\bp) & = \frac{1}{T}\sum_{t=1}^T \ell_{y_t}^{\tnr{val}}( \hat{z}_{t,\backslash{t}};\bp)\\
\label{hat.z.t.backlash.t}
    \hat{z}_{t,\backslash{t}} & = \bx\T_t\hat\btheta_{\backslash{t}}\\
\label{optimize.theta.i}
    % \hat\btheta_{\backslash{\tau}} &= \argmin_{\btheta}
    % \sum_{\substack{t=1\\ t\neq \tau}}^T
    % \ell^{\tnr{loss}}_{y_t}(\bx\T_t\btheta; \bp) + \rho(\btheta;\bp).
    \hat\btheta_{\backslash{t}} &= \argmin_{\btheta} \big[J(\btheta) - 
    \ell^{\tnr{loss}}_{y_{t}}(\bx\T_{t}\btheta; \bp)\big].
\end{align}
where $U(\bp)$ is the averaged validation metric.\footnote{Note that, while the \gls{loo} cross-validation uses the validation sets containing only one element, in general, we can use larger sets. However, they must then be defined arbitrarily (\eg randomly), as it is rarely possible to enumerate all of them. On the other hand, in the \gls{loo} approach we \emph{do} enumerate all $T$ validation sets. Thus, given the data, the results are independent of the random/arbitrary definition of the validation sets. This removes any ambiguity in the numerical optimization of $U(\bp)$ required in this work.}

We can also calculate the metrics for the matches played on the neutral and the home venues, given, respectively, by
\begin{align}
\label{U.ntr}
    U^{\tnr{ntr}}(\bp) & = \frac{1}{T^{\tnr{ntr}}}\sum_{\substack{t=1\\h_t=0}}^T \ell^{\tnr{val}}(\hat{z}_{t,\backslash{t}};\bp)\\
\label{U.hfa}
    U^{\tnr{hfa}}(\bp) & = \frac{1}{T^{\tnr{hfa}}}\sum_{\substack{t=1\\h_t=1}}^T \ell^{\tnr{val}}(\hat{z}_{t,\backslash{t}};\bp),
\end{align}
where, $U(\bp) = \big(T^{\tnr{ntr}}U^{\tnr{ntr}}(\bp) + T^{\tnr{hfa}}U^{\tnr{hfa}}(\bp)\big)/T$.

We note that  \eqref{U(p).define} can be transformed as follows:
\begin{align}
\label{V(p).define}
    V(\bp) = \e^{-U(\bp)}=\Big[\prod_{t=1}^T \mfP_{y_t}(\hat{z}_{t,\backslash{t}};\bp))\Big]^{\frac{1}{T}},
\end{align}
which is the geometric mean of the probabilities of observing $Y_t=y_t$, calculated from $\hat\btheta_{\backslash{t}}$. Thus, $V(\bp)$ is potentially easier to interpret than $U(\bp)$; and, with linearization, for small $U(\bp)$, \eg $U(\bp)<0.1$, we have $V(\bp)\approx 1- U(\bp)$. For example, $U(\bp)=1.4$ yields $V(\bp)\approx 24.7$\%, while for a uniform distribution $\mfP_{y}(z)=\frac{1}{6}, \forall y$ (\ie $V(\bp)=16.7$\%) we have $U(\bp)=1.79$. The values $V(\bp)$ are shown on the right-hand auxiliary axis in Figs.~\ref{fig:ALO_logloss}, Fig.~\ref{fig:ALO_FIVB}, and Fig.~\ref{fig:ALO_logloss_weights}. 

While the complexity is reduced, we still need to solve \eqref{optimize.theta.i} $T$ times, and, to alleviate the complexity, we apply here the \gls{alo} approach \citep{Beirami17},\citep{Rad20}, \citep{Burn20}, where \eqref{hat.theta} is solved once to find $\hat\btheta\equiv \hat\btheta(\bp)$ (this is where most of the computational complexity lies), and we make a quadratic approximation of the function $J(\btheta)$ around $\hat\btheta$, which allows us to find the closed-form approximation of \eqref{hat.z.t.backlash.t} as follows:
\begin{align}
\label{z.hat.t.backshlash.t}
	\hat{z}_{t,\backslash{t}} &\approx
    \hat z_t + 
    \frac{\dot\ell^{\tnr{loss}}_{y_t}(\hat{z}_t;\bp) a_t}{1-\ddot\ell^{\tnr{loss}}_{y_t}(\hat{z}_t;\bp)a_t},
\end{align}
where $a_t = \bx\T_t\hat{\matH}^{-1}\bx_t$, $\hat{\matH} = \nabla^2_{\btheta}J(\btheta)|_{\btheta=\hat\btheta}$ is the Hessian of the function $J(\btheta)$ under minimization in \eqref{hat.theta}, $\ddot\ell^{\tnr{loss}}_{y}(z;\bp)  = \frac{\dd}{\dd z}\dot\ell^{\tnr{loss}}_{y}(z;\bp)$, and $\hat{z}_t=\bx\T_t\hat\btheta(\bp)$. Details of the derivation can be found in \citep[Sec.~3]{Burn20} or \citep[Appendix.~2]{Szczecinski22a}. 

\subsection{Finding thresholds \boldmath{$c$} and \gls{hfa} parameter $\eta$}\label{Sec:Choose.bc}

% %%%%%%%%%%%%%%%%%%%%%%%%%%%%%%%%%%%%%%%%%%%%%%%%%%%
% \subsubsection{From empirical frequencies to the model}\label{Sec:freq.2.c.eta}

\begin{figure}[tb]
    \centering
    \includegraphics[width=1.0\linewidth]{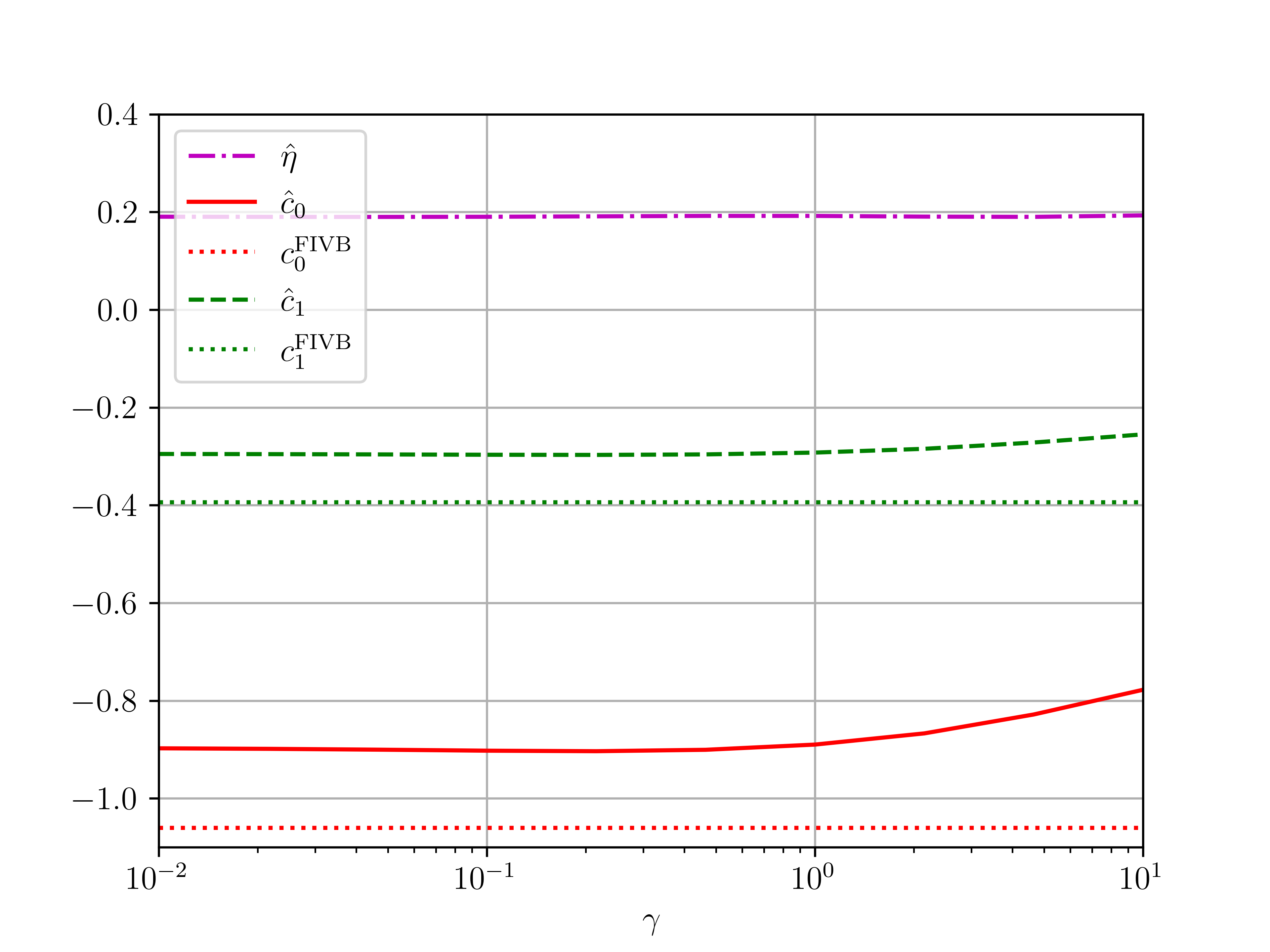}
    \caption{Results of optimization in \eqref{hat.bc.eta.logloss}: thresholds $\hat{c}_0$, $\hat{c}_1$ and the \gls{hfa} $\hat\eta$. The horizontal dotted lines are drawn for the values $c^{\tnr{FIVB}}_0=-1.06$ and $c^{\tnr{FIVB}}_1=-0.394$ used by the \gls{fivb} ranking, see \eqref{bc.FIVB}. All thresholds satisfy \eqref{c.l.are.symmetric} so $c_2\equiv 0$ need not be shown.}
    \label{fig:cy.eta}
\end{figure}

% %%%%%%%%%%%%%%%%%%%%%%%%%%%%%%%%%%%%%%%%%%%%%%%%%%%
% \subsubsection{Numerical optimization}\label{Sec:c.Numerical.optimization}
To assess the role of the thresholds $\bc^{\tnr{FIVB}}$ used in the \gls{cl} model \eqref{Qy(z).FIVB} and of the \gls{hfa} parameter $\eta$, we start with  $\ell^{\tnr{loss}}_y(z)=\ell_y(z)$ and consider four cases:
\begin{enumerate}
    \item[i)] We use $\bc=\bc^{\tnr{FIVB}}$ given by \eqref{bc.FIVB} and $\eta=0$; in other words, no optimization is performed.
    \item[ii)] 
    We find $\hat{\bc}$ by optimizing $U(\bp)$ in \eqref{optimize.hyper} but we set $\eta=0$, \ie
\begin{align}
\label{hat.bc.eta=0}
    \hat\bc =\argmin_{\bc, \eta=0} U(\bc)\quad \text{s.t.}\quad \bc~ \text{satisfies}~ \eqref{c.l.are.symmetric}.
\end{align}
\item[iii)] 
    We use $\bc^{\tnr{FIVB}}$ and optimize the \gls{hfa} parameter, \ie use
\begin{align}
\label{hat.eta.bc_FIVB}
    \hat\eta =\argmin_{\eta, \bc=\bc^{\tnr{FIVB}}} U(\eta);
\end{align}
    \item[iv)] 
    We find both $\hat{\bc}$ and $\hat\eta$ through optimization,
\begin{align}\label{hat.bc.eta.logloss}
    \hat\bc, \hat\eta =\argmin_{\bc,\eta} U(\bc,\eta)\quad \text{s.t.}\quad \bc~ \text{satisfies}~ \eqref{c.l.are.symmetric},
\end{align}
  and they are shown in Fig.~\ref{fig:cy.eta}.
\end{enumerate}

In the above, we slightly abuse the notation and write \eg $U(\bc)$ to indicate that we optimize only the parameter $\bc$ and all other hyperparameters in $\bp$ are kept constant (as we also explicitly indicate, when relevant, under the $\argmin$ operator); similarly, the notation $U(\bc,\eta)$ means that both $\bc$ and $\eta$ are optimized.

%%%%%%%%%%%%%%%%%%%%%%%%%%%%%%%%%%%%%%%%%%%%%%%%%%%%%%%%%%%%
%%%%%%%%%%%%%%%%%%%%%%%%%%%%%%%%%%%%%%%%%%%%%%%%%%%%%%%%%%%%

\begin{figure}[tb]
    \centering
    \includegraphics[width=1.0\linewidth]{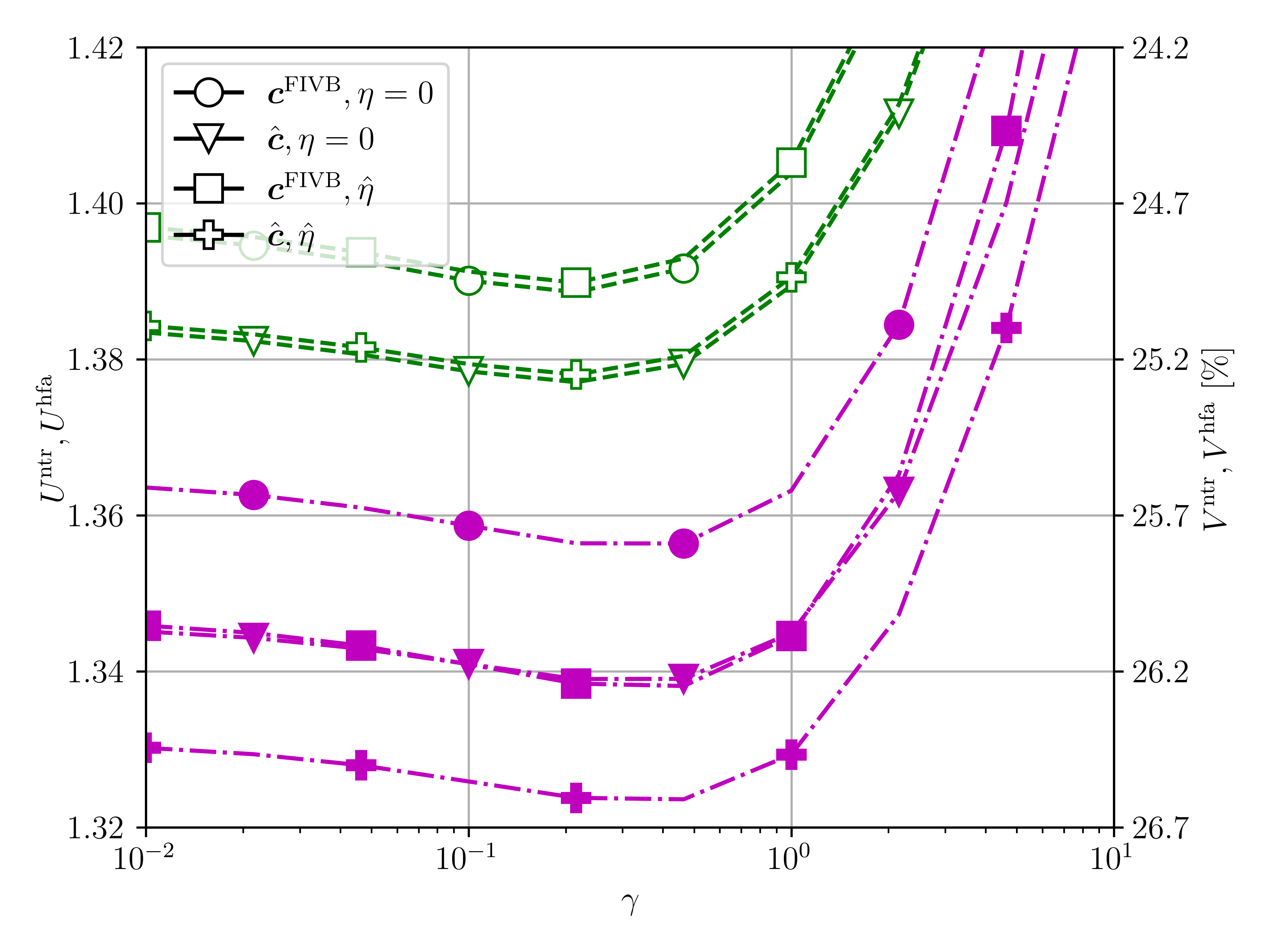}
    \caption{Validation metrics $U^{\tnr{ntr}}$ (hollow markers) and $U^{\tnr{hfa}}$ (colored markers) given by \eqref{U.ntr}-\eqref{U.hfa} shown as a function of $\gamma$ using the loss function $\ell_y(z)$ defined in \eqref{log.score}.}
    \label{fig:ALO_logloss}
\end{figure}

We show, in Fig.~\ref{fig:ALO_logloss}, cross-validation results $U^{\tnr{ntr}}(\bp)$ and $U^{\tnr{hfa}}(\bp)$ defined in \eqref{U.ntr} and \eqref{U.hfa} as functions of the regularization parameter $\gamma$, where $\bp$ contains all parameters, including $\gamma$ and others that are optimized according to the cases we explain above. The right-hand axis in Fig.~\ref{fig:ALO_logloss} shows the metric $V(\bp)$ as a transformation of $U(\bp)$ through \eqref{V(p).define}.

%%%%%%%%%%%%%%%%%%%%%%%%%%%%%%%%%%%%%%%%%%%%%%%%%%%%%%%%%%%%
%%%%%%%%%%%%%%%%%%%%%%%%%%%%%%%%%%%%%%%%%%%%%%%%%%%%%%%%%%%%

~\\\noindent
\textbf{Main observations}

From Fig.~\ref{fig:cy.eta}, we see that the \gls{fivb} thresholds $\bc^{\tnr{FIVB}}$ are close to, but not identical with those obtained through optimization. This discrepancy should be attributed to the difference in the data sets from which the thresholds were inferred, but also to the procedure described by the \gls{fivb} ranking, which defined $\bc^{\tnr{FIVB}}$ from the matches of the ``teams with similar skills".\footnote{We note that this approach to find the model is rather ambiguous as, to find which skills are similar, we have to estimate them, and this requires the model to be defined in the first place.} In our approach, we used all matches, which may explain some improvement in the validation metrics $U(\hat\bp)$ we observe in Fig.~\ref{fig:ALO_logloss}. 

From Fig.~\ref{fig:ALO_logloss} we  observe that:
\begin{itemize}
    \item The decomposition into the matches played on the neutral and home venues indicates that, by including the \gls{hfa} into the model via $\eta$, we can improve the prediction for the home matches, and, rather unsurprisingly, this modification has practically no impact on the neutral-venue matches.

    \item Attention should be paid to the metric $V(\bp)$ shown on the right axis, where we see that the improvements are on the order of a fraction of a percent. For example, the most significant improvement appears for home matches, where, by optimizing $\eta$ and the thresholds $\bc$ in \eqref{hat.bc.eta.logloss}, the results are improved by $\sim 1\%$; \ie from $V\approx25.8\%$ to $V\approx26.8\%$. %This can be compared to the performance of the empirical entropy shown in Table~\ref{Table:Entropy.values}, which is notably worse. 
\end{itemize}

%%%%%%%%%%%%%%%%%%%%%%%%%%%%%%%%%%%%%%%%%%%%%%%%%%%%%%%%%%%%%%%%%%%%%%
\subsection{Finding numerical score \boldmath $r$}\label{Sec:Numerical.score}

To discuss the role of the numerical scores $\br$ used in the \gls{fivb} algorithm, we will first, in \secref{Sec:From.c.to.r}, analyze the implicit loss functions used by the algorithm while a purely numerical analysis / optimization is carried out in \secref{Sec:Numerical.optimization.of.r}.

%%%%%%%%%%%%%%%%%%%%%%%%%%%%%%%%%%%%%%%%%%
\subsubsection{From thresholds to numerical scores}\label{Sec:From.c.to.r}

\begin{figure}[tb]
    \centering
    \includegraphics[width=1.0\linewidth]{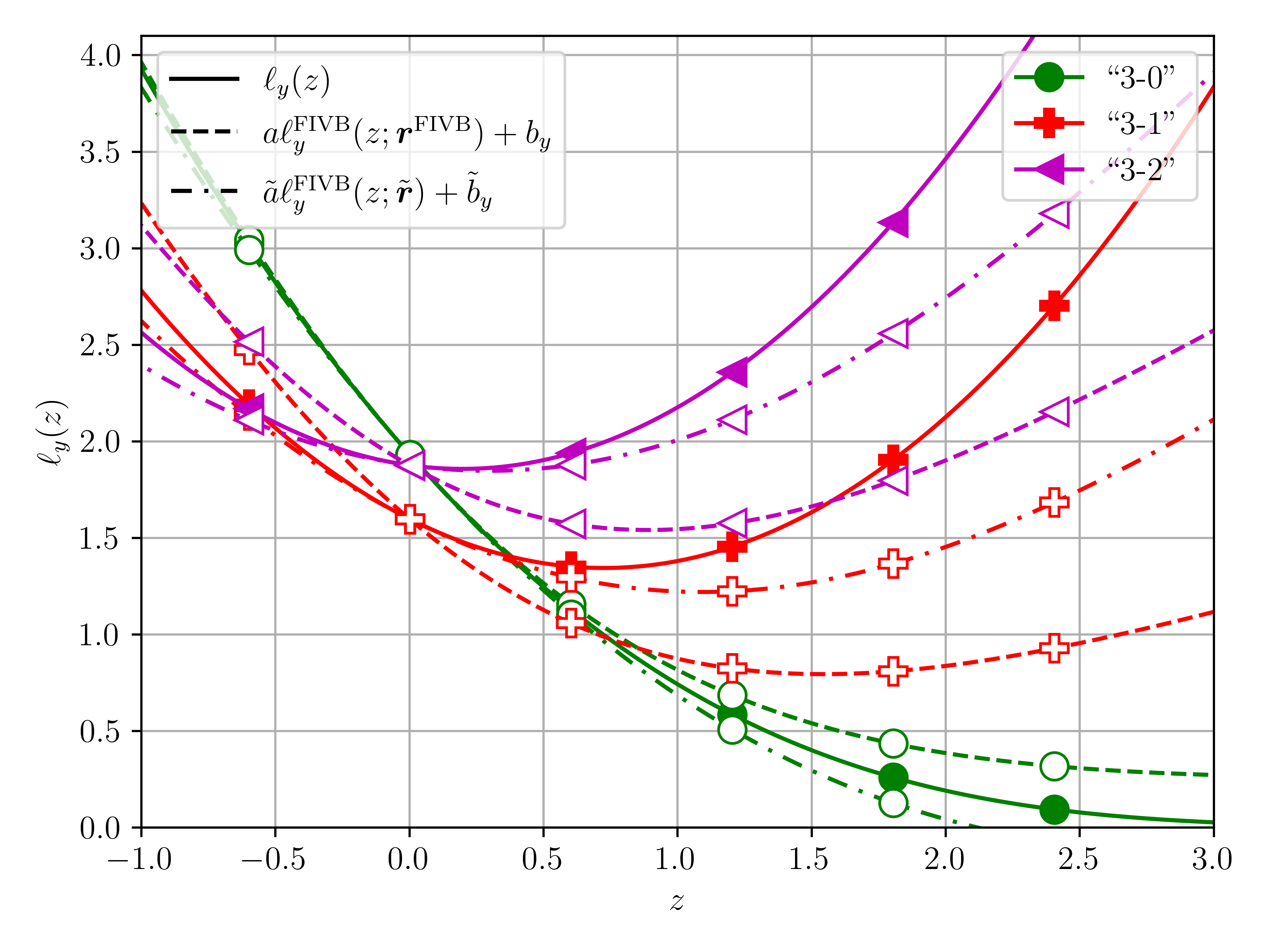}
    \caption{Loss functions: log-score $\ell_y(z)$ (solid line) and rescaled/shifted  implicit loss function $a\ell^{\tnr{FIVB}}_y(z;\br)+b_y$ (dashed line) for $y=0,1,2$ corresponding, respectively, to the outcomes ``3-0", ``3-1", and ``3-2". %The dashed-dotted line corresponds to $\tilde{a}\ell^{\tnr{FIVB}}_y(z;\tilde\br)+\tilde{b}_y$ calculated using $\tilde{\br}$ given in \eqref{tilde.r.for.FIVB}-\eqref{tilde.r.for.FIVB.2}. The loss functions for $y=3, 4, 5$ can be obtained through symmetry $\ell_y(z)=\ell_{L-1-y}(-z)$.
    }
    \label{fig:ell.FIVB}
\end{figure}

We show, in Fig.~\ref{fig:ell.FIVB}, the logarithmic loss functions $\ell_y(z)$, as well as, the scaled and vertically-shifted versions of the implicit loss functions $a\ell^{\tnr{FIVB}}_y(z;\br)+ b_y$, where we find $a$ by matching the first derivatives of the loss function $\ell_0(z_{\tnr{o}})$ for $z=z_{\tnr{o}}=0$
\begin{align}\label{dot.ell.0=a.g.FIVB}
    \dot\ell_0(z_{\tnr{o}})&=a g^{\tnr{FIVB}}_0(z_{\tnr{o}};\br),
\end{align}
which, from $g^{\tnr{FIVB}}_y(z_{\tnr{o}};\br)=\check{r}(z_{\tnr{o}})-r_y=-r_y$,\footnote{Because, due to symmetry, the expected numerical score is zero, \ie $\check{r}(0)=0$} yields 
\begin{align}
    \label{alpha.from.dot.ell.r_0}
    a&=-\frac{\dot\ell_0(z_{\tnr{o}})}{r_0}.
\end{align}
Similarly, the shifts $b_y$ are calculated to match the values of the loss functions at $z_{\tnr{o}}$, \ie to satisfy $\ell_y(z_{\tnr{o}})=a\ell^{\tnr{FIVB}}_y(z_{\tnr{o}};\br)+b_y$. This means that $b_y=\ell_y(z_{\tnr{o}})-a\ell^{\tnr{FIVB}}_y(z_{\tnr{o}};\br)$. 

This scaling/shifting transformation is irrelevant from the optimization point of view\footnote{Scaling \emph{all} loss function with $a>0$, obviously, does not change the results of \eqref{hat.theta.map.sum}. Similarly, adding $b_y$ to each loss function does not affect optimality}, but allows us to visually appreciate the difference between the loss functions in the vicinity of the target value $z_{\tnr{o}}=0$. Note that assuming that $z_t$ will be mostly observed close to $z_{\tnr{o}}$ is compatible with $z$ being a zero-mean Gaussian variable, which is the modeling assumption we used in Sec.~\ref{Sec:ranking=inference}.

%%%%%%%%%%%%%%%%%%%%%%%%%%%%%%%%%%%%%%%%%%%%%%%%%%%
%%%%%%%%%%%%%%%%%%%%%%%%%%%%%%%%%%%%%%%%%%%%%%%%%%%

Indeed, Fig.~\ref{fig:ell.FIVB} indicates that an almost perfect match is obtained for $\ell_0(z)$, \ie the logarithmic loss is practically indistinguishable from the scaled/shifted implicit loss. However, the loss functions are not matched for $y=1, 2$, and we want to find the numerical score $\tilde\br=[\tilde{r}_0,\ld,\tilde{r}_{L-1}]$, which satisfies a generalized version of \eqref{dot.ell.0=a.g.FIVB}, \ie we want the latter to hold for all $y\in\mcY$, that is
\begin{align}\label{dot.ell=g.FIVB.at.zero}
    \dot\ell_y(z_{\tnr{o}}) = a g_y^{\tnr{FIVB}}(z_{\tnr{o}};\tilde\br),\quad y\in\mcY.
\end{align}
%Note that, for the purpose of the discussion, $\br$ are shown as parameters of $g^{\tnr{FIVB}}_y(z;\br)$ and thus, are also parameters of the implicit loss function  $\ell^{\tnr{FIVB}}_y(z;\br)$.

Since $\dot\ell_y(z_0)=g_y^{\tnr{FIVB}}(z_{\tnr{o}};\tilde\br) = -\tilde{r}_y$, \eqref{dot.ell=g.FIVB.at.zero} is solved by
\begin{align}\label{r.y.simple}
    \tilde{r}_y\equiv\tilde{r}_y(\bc) &=  -\tilde{r}_0\frac{\dot\ell_y(z_{\tnr{o}})}{\dot\ell_0(z_{\tnr{o}})} = \tilde{r}_0\frac{\Phi(c_0)\big(\mcN(c_y)-\mcN(c_{y-1})\big)}{\mcN(c_0)\big(\Phi(c_y)-\Phi(c_{y-1})\big)},\quad y\in\mcY,
\end{align}
where we used \eqref{dot.log.score.FIVB} and \eqref{alpha.from.dot.ell.r_0}; the notation $\tilde{r}_y(\bc)$ emphasizes the dependence of $\tilde{\br}$ on the thresholds $\bc$ which define the \gls{cl} model so \eqref{r.y.simple} is valid for any $\bc$.  Note also that $\tilde{r}_y$ is not uniquely defined because we can arbitrarily fix  $\tilde{r}_0$, and, for comparison with the \gls{fivb} ranking, we set $\tilde{r}_0\equiv r^{\tnr{FIVB}}_0=2.0$.

We can now calculate the scores by applying \eqref{r.y.simple} to the \gls{fivb}-defined thresholds, $\bc=\bc^{\tnr{FIVB}}$, which yields $\tilde\br=[\tilde{r}_0(\bc^{\tnr{FIVB}}), \ld,\tilde{r}_{L-1}(\bc^{\tnr{FIVB}})]$
\begin{align}
\label{tilde.r.for.FIVB}
    \tilde{r}_0&=-\tilde{r}_{5}\equiv2.0,\\ 
\label{tilde.r.for.FIVB.1}
    \tilde{r}_1(\bc^{\tnr{FIVB}})&=-\tilde{r}_{4}(\bc^{\tnr{FIVB}})=0.89,\\ 
\label{tilde.r.for.FIVB.2}
    \tilde{r}_2(\bc^{\tnr{FIVB}})&=-\tilde{r}_{3}(\bc^{\tnr{FIVB}})=0.25.
\end{align}

These values are rather different from those shown in Table~\ref{tab:numerical.values.FIVB} and, more importantly, when we use them to calculate the implicit loss functions $\ell^{\tnr{FIVB}}_y(z;\tilde\br), y=0,1,2$ (shown, scaled with $\tilde{a}$ and shifted with $\tilde{b}_y$, in Fig.~\ref{fig:ell.FIVB} with dashed-dotted lines), the latter are indistinguishable from $\ell_y(z)$ in the vicinity of $z_{\tnr{o}}=0$. Clearly, when compared to the implicit loss $\ell_y^{\tnr{FIVB}}(z;\br^{\tnr{FIVB}})$ with \gls{fivb}-defined scores $\br^{\tnr{FIVB}}$, the loss $\ell^\tnr{FIVB}_y(z;\tilde\br)$ offers a better approximation of the log-loss $\ell_y(z)$. And this improvement is obtained solely using the numerical score $\tilde\br$.
%We note that, instead of matching the derivatives at $z_{\tnr{o}}=0$, we may opt for a different criterion. For example, we might set $z_{\tnr{o}}\in(0,\eta)$ to take into account the shift of the argument in the matches played at the home venues, see \eqref{hfa.loss.definition}. More generally, the fit can be weighted to ensure that the gradients of the loss functions are ``well" approximated, not only for $z=z_{\tnr{o}}$ but also in the larger range of values,
%\footnote{For example, we might solve:
% \begin{align}
%     \hat\br &= \argmin_{\br} d(\br),\\
%     d(\br)=&\sum_{y=0}^{L-1} \tilde{k}^{\tnr{ntr}}_y \int |\dot\ell_y(z)-g^{\tnr{FIVB}}_y(z)|^2 \mcN(z;0,2v_{\theta})\dd z\\
%     &+
%     \sum_{y=0}^{L-1} k^{\tnr{hfa}}_y \int |\dot\ell_y(z+\eta)-g^{\tnr{FIVB}}_y(z+\eta)|^2 \mcN(z;0,2v_{\theta})\dd z,
% \end{align}
% } 
%but the simplicity of \eqref{r.y.simple} likely trumps any more sophisticated, and, thus also complex, approaches.

%%%%%%%%%%%%%%%%%%%%%%%%%%%%%%%%%%%%%%%%%%%%%%%%%%%%%%%%%%%%
\subsubsection{Numerical optimization}\label{Sec:Numerical.optimization.of.r}

Instead of the analytical approach, shown in the previous section, the numerical optimization of $\br$, takes into account the actual outcomes of the matches. 

To analyze the optimality of $\br^{\tnr{FIVB}}$, the loss function $\ell^{\tnr{loss}}_y(z)$ is set to $\ell^{\tnr{FIVB}}_y(z)$, and we consider the following cases:
\begin{enumerate}
    \item[i)] We use $\bc^\tnr{FIVB}$ and $\br^\tnr{FIVB}$ specified by the \gls{fivb} ranking, and given, respectively, in \eqref{bc.FIVB} and Table~\ref{tab:numerical.values.FIVB}, and we set $\eta=0$. This is the reference, currently used by \gls{fivb}.
    \item[ii)] 
    We optimize $\br$, $\eta$ for a given $\bc$
    \begin{align}\label{hat.br}
        \hat\br(\bc), \hat\eta(\bc) =\argmin_{\br,\eta} U(\br,\eta,\bc,\bp_{\backslash\set{\br,\eta,\bc}}),
    \end{align}
    where we consider $\bc=\bc^\tnr{FIVB}$ and $\bc=\hat\bc$, with the latter obtained via \eqref{hat.bc.eta.logloss}.
    \item[iii)] 
    We \emph{calculate} the numerical score $\tilde\br=\tilde\br\big(\bc\big)$ using the formula shown in \eqref{r.y.simple}, and set $\eta=0.2$ (which is a rounded value of $\hat\eta$ obtained through optimization; see Fig.~\ref{fig:cy.eta}). As before, this is done for $\bc=\bc^\tnr{FIVB}$ and $\bc=\hat\bc$.
\end{enumerate}

%%%%%%%%%%%%%%%%%%%%%%%%%%%%%%%%%%%%%%%%%%%%%%%%%%%%%%%%%%%%
\begin{figure}[tb]
    \centering
    \includegraphics[width=1.0\linewidth]{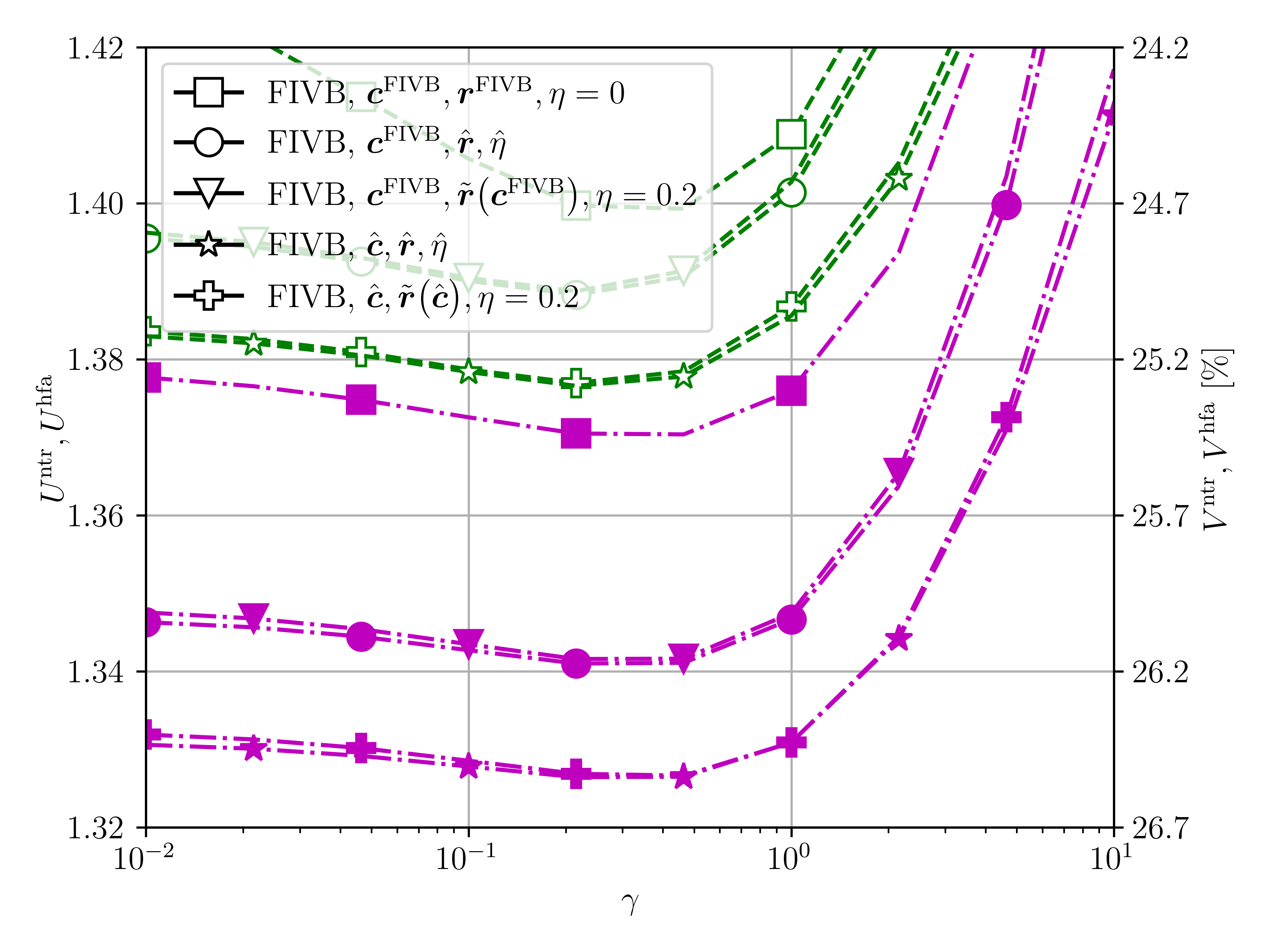}
    \caption{Validation metrics $U^{\tnr{ntr}}$ (hollow markers) and $U^{\tnr{hfa}}$ (colored markers) given by \eqref{U.ntr}-\eqref{U.hfa} shown as a function of $\gamma$ for
    the loss function $\ell^{\tnr{loss}}_y(z)=\ell^{\tnr{FIVB}}_y(z)$ defined in \eqref{implicit.loss.FIVB} using parameters which may be: predefined ($\br^{\tnr{FIVB}}$), analyticaly calculated ($\tilde{\br}\big(\bc^{\tnr{FIVB}}\big)$), or, numerically optimized ($\hat{\br}$).}
    \label{fig:ALO_FIVB}
\end{figure}
%%%%%%%%%%%%%%%%%%%%%%%%%%%%%%%%%%%%%%%%%%%%%%%%%%%%%%%%%%%%

The \gls{alo} metrics are shown in Fig.~\ref{fig:ALO_FIVB}, where we observe:
\begin{itemize}
    \item The ALO metric obtained with optimized numerical scores $\hat\br$ and with the calculated ones $\tilde\br(\bc)$ are practically identical, which supports our analysis in Sec.~\ref{Sec:From.c.to.r}. 
        
    \item By comparing the results from Fig.~\ref{fig:ALO_FIVB} with those shown in Fig.~\ref{fig:ell.FIVB} we see that, using the implicit \gls{fivb} loss functions, $\ell^{\tnr{FIVB}}_y(z)$, and provided the numerical scores $\br$ are adequately set (\ie optimized via \eqref{hat.br} or calculated via \eqref{r.y.simple}), a negligible loss of performance is incurred when comparing to the log-score $\ell_y(z)$. This justifies the choice made in the  \gls{fivb} ranking which avoids the exact, but complicated derivative of the loss function shown in \eqref{dot.log.score.FIVB}.

    \item The numerical scores $\br^{\tnr{FIVB}}$ currently used in the \gls{fivb} ranking, lead to the observable performance loss.       
\end{itemize}

\begin{figure}[tb]
    \centering
    \includegraphics[width=1.0\linewidth]{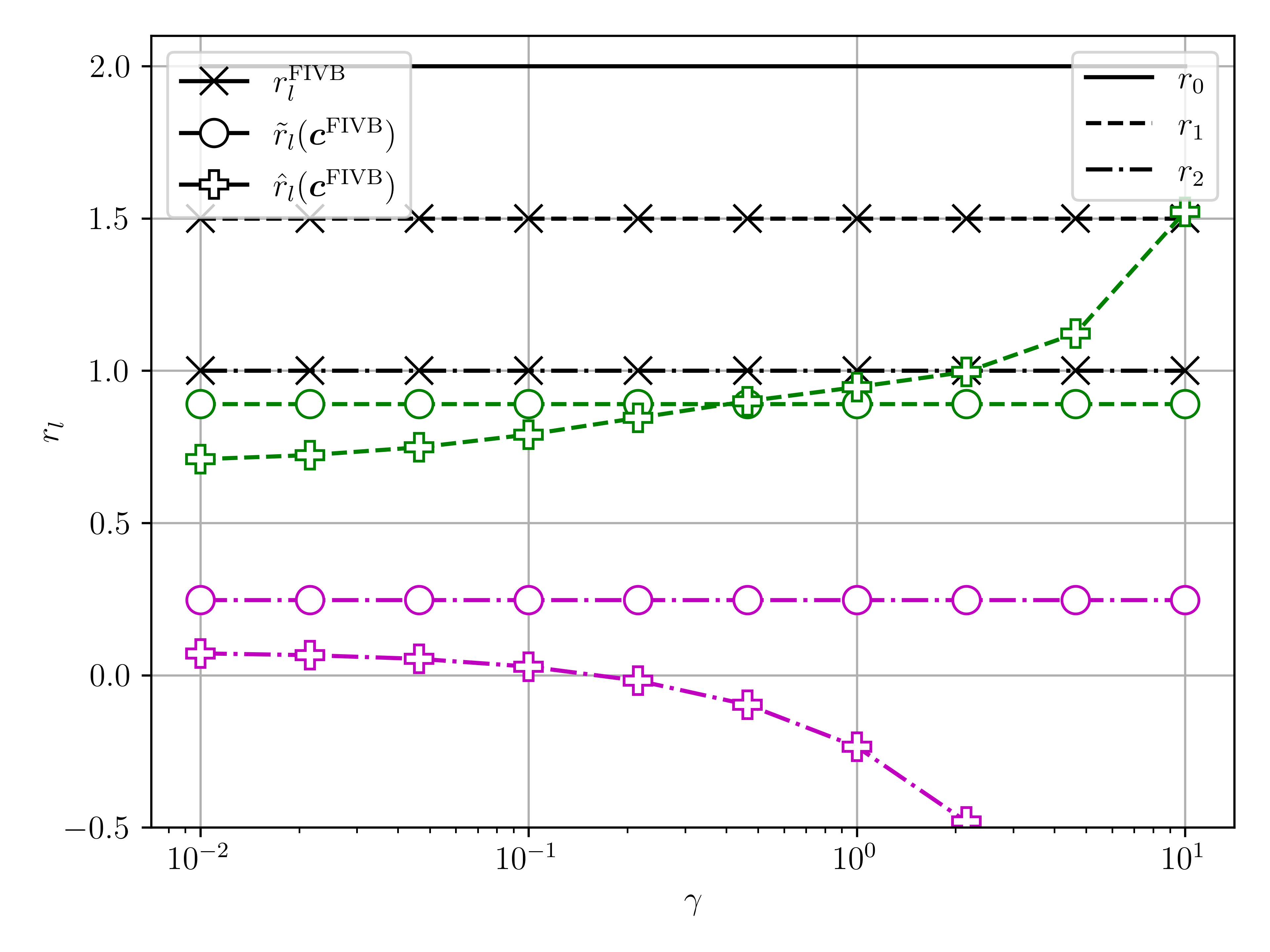}
    \caption{Numerical scores: $r_l^{\tnr{FIVB}}$ given by the \gls{fivb} ranking, $\tilde{r}_l(\bc^{\tnr{FIVB}})$ calculated via \eqref{r.y.simple}, and $\hat{r}_l(\bc^{\tnr{FIVB}})$ optimized in \eqref{hat.br}.}
    \label{fig:numerical.score}
\end{figure}

The numerical score $\br^{\tnr{FIVB}}$ used currently in the ranking is compared, in Fig.~\ref{fig:numerical.score}, to $\hat\br(\bc^{\tnr{FIVB}})$ which is obtained by optimization \eqref{hat.br} and to $\tilde\br(\bc^{\tnr{FIVB}})$ -- calculated from \eqref{r.y.simple}; since $\br^{\tnr{FIVB}}$ and $\tilde\br(\bc^{\tnr{FIVB}})$ do not depend on the regularization parameter $\gamma$, they are shown as horizontal lines. Also note that $r_0^{\tnr{FIVB}} = \tilde{r}_0(\bc^{\tnr{FIVB}}) =  \hat{r}_0(\bc^{\tnr{FIVB}}) = 2$.

The optimized numerical scores $\hat\br$ change with $\gamma$, but have no incidence on the prediction capacity of the model, as shown already in Fig.~\ref{fig:ALO_FIVB}. In fact, for $\gamma\approx 0.5$, which minimizes the total loss function, we get $\tilde{r}_1(\bc^{\tnr{FIVB}})\approx \hat{r}_1(\bc^{\tnr{FIVB}})$.

%%%%%%%%%%%%%%%%%%%%%%%%%%%%%%%%%%%%%%%%%%%%%%%%%%%%%%%%%%%%
%%%%%%%%%%%%%%%%%%%%%%%%%%%%%%%%%%%%%%%%%%%%%%%%%%%%%%%%%%%%

On the other hand, $\hat{r}_2(\bc^{\tnr{FIVB}})$ becomes negative. For example, using the results for $\gamma\approx 0.5$ implies using, 
\begin{align}
\label{hat.r}
    \hat{r}_0&=2.0,\quad 
    \hat{r}_1\approx0.9,\quad 
    \hat{r}_2\approx -0.1,\quad
    \hat{r}_3\approx 0.1,\quad
    \hat{r}_4\approx-0.9,\quad
    \hat{r}_4=-2,
\end{align}
\ie the numerical score does not decrease monotonically with the outcomes index, $y$.

This may appear surprising and counterintuitive, but only if we interpret the numerical score as related to the order of the outcomes. We should remember that ordinal variables do not have intrinsic numerical values, and numerical scores are parameters that allow us to adjust the form of the implicit loss function $\ell^{\text{FIVB}}_y(z)$. With such a perspective, the non-monotonic behavior of $r_y$ is allowed.\footnote{
We still want to know, if, for $\hat\br$ given in \eqref{hat.r} (where $\hat{r}_3>\hat{r}_2$) the implicit loss functions $\ell^{\tnr{FIVB}}_y(z;\hat\br)$ remain convex in $z$. For this, it suffices to calculate $\check{r}(z)$ and verify that it is monotonically increasing in $z$. In fact, this is the case for all $\hat\br$ we obtained. Note that this does not contradict Lemma~\ref{lemma:1} because the monotonic behavior of $r_y$ was a sufficient (but not necessary) condition to ensure the convexity of $\ell^{\tnr{FIVB}}_y(z; \br)$.}

We also note that we always obtain more ``conventional'' (monotonic in $y$) behavior of $\tilde{r}_y(\bc)$. Since the optimized scores, $\hat\br$, and the calculated ones, $\tilde\br$, do not change the performance, it may be preferable to use the latter.

Immediate conclusion is that, the numerical score $\br^{\tnr{FIVB}}$ is inadequately set in the current version of the \gls{fivb} ranking. However, it can be easily modified, \eg using rounded values, $\tilde{r}_1=1.0$ and $\tilde{r}_2=0.25$.

%This conclusion is different in nature from our observation about the optimality of the thresholds $\bc^{\tnr{FIVB}}$. Not only are they quite similar to the values $\hat{\bc}$ we found, but we may argue that the discrepancy is due to differences between the datasets used by us and the designers of the \gls{fivb} ranking. 

%On the other hand, the numerical scores $\br^{\tnr{FIVB}}$ do not correspond to the optimal solution, even if we use $\bc^{\tnr{FIVB}}$. This conclusion is corroborated by the numerical and analytical results we just presented.

Thus, it appears that the numerical scores $\br^{\tnr{FIVB}}$ were not formally optimized --- a conjecture supported by the fact that their origin is not explained in \citep{fivb_rating}. However, regardless of the origin of $\br^{\tnr{FIVB}}$, it is more sound to see the numerical score $\br$ as free parameters which allow us to make the implicit loss function $\ell^{\tnr{FIVB}}_y(z;\br)$ ``behave" similarly to the optimal logarithmic loss $\ell_y(z)$.

%We \emph{speculate} that the numerical scores $\br$ are defined to be similar to the points in the volleyball leagues, where winning ``$3-0$" grants $3$ points to the winner, winning ``$3-1$" grants $2$ points, and winning ``$3-2$" grants $1$ point. Multiplying these points by $2/3$, this scoring rule would be equivalent to granting $2.0$, $1.33$, and $0.66$ points -- values somewhat similar to $r^{\tnr{FIVB}}_0=2.0$, $r^{\tnr{FIVB}}_1=1.5$, and $r^{\tnr{FIVB}}_2=1.0$. 

%%%%%%%%%%%%%%%%%%%%%%%%%%%%%%%%%%%%%%%%%%%%%%%%%%%%%%%%%%%%
\begin{figure}[tb]
    \centering
    \includegraphics[width=1.0\linewidth]{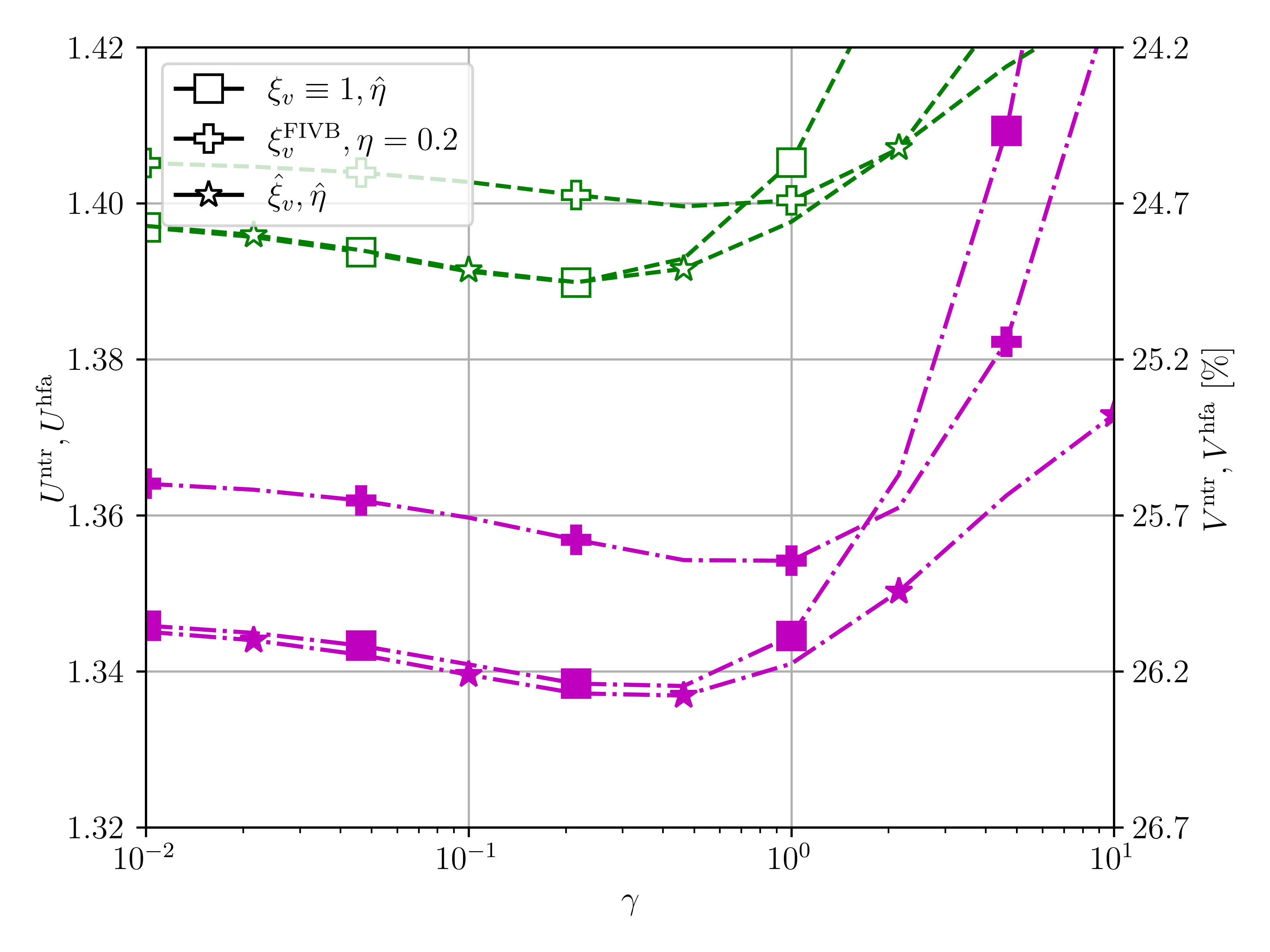}
    \caption{Validation metrics $U^{\tnr{ntr}}$ (hollow markers) and $U^{\tnr{hfa}}$ (colored markers) given by \eqref{U.ntr}-\eqref{U.hfa} shown as a function of $\gamma$ using the loss function $\ell_y(z)$ defined in \eqref{weighted.loss.definition} and different strategies of fixing the weights $\xi_v$ which depend on the matches' categories, including equal weights $\xi_v\equiv1$,  $\xi^{\tnr{FIVB}}_v$ specified in Table~\ref{tab:FIVB.xi}, and $\hat\xi_v$, optimized via \eqref{hat.bxi}.}
    \label{fig:ALO_logloss_weights}
\end{figure}
%%%%%%%%%%%%%%%%%%%%%%%%%%%%%%%%%%%%%%%%%%%%%%%%%%%%%%%%%%%%

%%%%%%%%%%%%%%%%%%%%%%%%%%%%%%%%%%%%%%%%%%%%%%%%%%%%%%%%%%%%%%%%%%%%%%
\subsection{Weights}\label{Sec:categ.weighting}

The weights $\bxi$ are optimized as follows:
\begin{align}\label{hat.bxi}
    \hat\bxi, \hat\eta =\argmin_{\br,\eta} U(\br,\eta, \bp_{\backslash\set{\br,\eta}}),
\end{align}
where we use the thresholds $\bc^{\tnr{FIVB}}$ and the log-score function $\ell^{\tnr{loss}}_y(z)=\ell_y(z)$. The starting point for optimization is $\bxi=[1, \ld, 1]$.

The results are shown in Fig.~\ref{fig:ALO_logloss_weights}, and the conclusion is straightforward: using the weights $\xi_v$ specified by the \gls{fivb} ranking and given in Table~\ref{tab:FIVB.xi}, is detrimental to the prediction capacity of the model. The optimization is also practically useless, and, in fact, the optimized weights were quite similar. That is, for $\gamma<0.5$, we obtained $\hat\xi_v \in (0.9, 1.5)$ (not shown here).

To clarify the somewhat intriguing behavior of $U(\bp)$ for the optimized results (marked with stars in Fig.~\ref{fig:ALO_logloss_weights}), which, for large $\gamma$, is notably better than in the case of $\xi_v\equiv 1$, we write the optimization \eqref{optimize.theta.i} as
\begin{align}
    \hat\btheta_{\backslash{t}} 
    &=\argmin_{\btheta} \Big[\sum_{\substack{\tau=1\\\tau\neq t}}^T \hat\xi_{v_{\tau}}(\hat\gamma)\ell^{\tnr{loss}}_{y_{\tau}}(\bx_{\tau}\T\btheta) + \hat\gamma \|\btheta\|^2\Big].
\end{align}
where $\hat\xi_v(\hat\gamma)$ are the optimal weights obtained in \eqref{hat.bxi} for an optimal $\hat\gamma\approx 0.5$. Since the multiplication of the cost function by $\alpha>0$ is irrelevant to the optimization results, using weights $\alpha\hat\xi_v(\gamma)$ and the regularization parameter $\gamma=\alpha\hat\gamma$, will not change $\hat\btheta_{\backslash{t}}$. In other words, increasing $\gamma$ we can simultaneously increase $\bxi$ and maintain performance $U(\bp)$ flat in $\gamma$. The only reason it does not happen is because we impose the constraint $\xi_0=1$.

\section{Real-time ranking}\label{Sec:Real.time.ranking}

In previous sections, model evaluation relied on comparing analytically deduced parameters with those obtained through optimization. Our objective now is to directly use the model obtained thanks to analytical insights and to evaluate the performance of the real-time ranking based on the resulting \gls{sg} algorithm. 

We will thus keep the \gls{fivb} model defined by the threshold parameters $\bc^{\tnr{FIVB}}$ and evaluate i) the choice of the numerical score $\br$ used in the implicit loss function $\ell^{\tnr{FIVB}}$, and specified in \eqref{tilde.r.for.FIVB} and  ii) the values of the weights $\bxi$, see Sec.~\ref{Sec:categ.weighting}, 
%and iii) the choice of the threshold parameters in the logistic \gls{cl} model, see Sec.~\ref{Sec:CL.logistic.model}. 
Simply put, we do not carry out any explicit optimization of the model parameters when using the real-time ranking but, rather, rely on the parameters obtained from analysis. In this way, we avoid the contentious issue of choosing the model parameters from the data. The only exception is the choice of the \gls{hfa} parameter which we set as $\eta=0.2$. %However, even in this case, this result is deduced from the frequencies of the outcomes and thus, does not rquire the explicit optimization of the skills.

To initialize the \gls{sg} algorithm \eqref{SG.btheta}, we use the skills $\theta_{m,0}, m=1,\ld, M$, where $\theta_{m,0}$ is read from the official \gls{fivb} ranking of the team $m$ at the time of their first match after 2020.\footnote{Thus, we do not take into account the fact that the \gls{fivb} ranking penalizes ``inactive'' teams, \ie those that do not play any matches in a given year, and whose ranking is then reduced by 50 points.} By initializing the skills with those provided by the official ranking allows us to deal with the practical aspect of switching from one ranking (here, the official one) to another (the one we propose).

% In order to switch the model from the Gaussian to the logistic \gls{cdf}, we need to change the scaling, but this is also handled analytically: to obtain a quasi-equivalence of the model, the logistic model \eqref{Q.y.z.CL.logistic} needs to employ scaling $s^{\mcL}=\sigma s\approx 80$ (where $s=125$, see \eqref{Qy(z).FIVB}) and use $\eta^{\mcL}=s/\sigma\approx 0.12$. Note that a similar strategy for adjusting the scales to avoid convergence problems is also discussed in \citep[Sec.~5.1]{Szczecinski22a}, but it is based on numerical results, while we do it analytically.

We calculate the validation metrics
\begin{align}
\label{U.SG}
    \ov{U}
    &= \frac{1}{T}\sum_{t=1}^{T}
    \ell^{\tnr{val}}_{y_t}(\bx_t\T\hat\btheta_{t})\\
\label{U.ntr-SG}
    \ov{U}^{\tnr{ntr}}
    &= \frac{1}{T^{\tnr{ntr}}}\sum_{\substack{t=1\\h_t=0}}^{T}
    \ell^{\tnr{val}}_{y_t}(\bx_t\T\hat\btheta_{t})\\
\label{U.hfa-SG}
    \ov{U}^{\tnr{hfa}}
    &= \frac{1}{T^{\tnr{hfa}}}\sum_{\substack{t=1\\h_t=1}}^{T}
    \ell^{\tnr{val}}_{y_t}(\bx_t\T\hat\btheta_{t}).
\end{align}
These metrics are, in essence, equivalent to those in \eqref{U.ntr}-\eqref{U.hfa}, which we have shown in the figures. The difference is that now skills $\hat\btheta_t$ are estimated using the \gls{sg} algorithm.

The results obtained are shown in Table~\ref{tab:SG.results}, where we indicate the loss function used (which determines the gradient used in the \gls{sg} algorithm) and the parameters of the underlying model.

%%%%%%%%%%%%%%%%%%%%%%%%%%%%%%%%%%%
\begin{table}[tb]
    \centering
    \begin{tabular}{c|c|l|c|ccc|c}
    & loss & parameters & $\mu$ & $\ov{U}$ & $\ov{U}^{\tnr{ntr}}$ & $\ov{U}^{\tnr{hfa}}$ & $\ov{\rho}$\\
    \hline\hline
    \multirow{2}{*}{A}  &
    \multirow{2}{*}{$\ell^{\tnr{FIVB}}$}  & 
    \multirow{2}{*}{ $\bc^{\tnr{FIVB}}, \br^{\tnr{FIVB}}, \bxi^{\tnr{FIVB}}, \eta=0$}    
        & 0.01 & 1.52 & 1.51 & 1.53 & 0.94\\
    &&  & 0.03 & 1.49 & 1.49 & 1.49 & 0.89\\
    \hline
    \multirow{2}{*}{B}  &
    \multirow{2}{*}{$\ell^{\tnr{FIVB}}$}  & \multirow{2}{*}{ $\bc^{\tnr{FIVB}}, \br^{\tnr{FIVB}}, \bxi^{\tnr{FIVB}}, \eta=0.2$}  
        & 0.01 & 1.52 & 1.51 & 1.52 & 0.94\\
    &&   & 0.03 & 1.48 & 1.49 & 1.47 & 0.89\\
    \hline
    \multirow{2}{*}{C}  &
    \multirow{2}{*}{$\ell^{\tnr{FIVB}}$}  & \multirow{2}{*}{ $\bc^{\tnr{FIVB}}, \tilde\br, \bxi^{\tnr{FIVB}}, \eta=0.2$}
        & 0.01 & 1.52 & 1.51 & 1.53 & 0.94\\
    &&   & 0.04 & 1.47 & 1.48 & 1.45 & 0.88\\
    \hline
    D  &
    \rule{0pt}{2.5ex}
    $\ell^{\tnr{FIVB}}$  & $\bc^{\tnr{FIVB}}, \br^{\tnr{FIVB}}, \xi_v\equiv 1, \eta=0.2$                   & 0.10 & 1.48 & 1.49 & 1.45 & 0.87\\
    \hline
    E  &
    \rule{0pt}{2.5ex}
    $\ell^{\tnr{FIVB}}$  & $\bc^{\tnr{FIVB}}, \tilde\br, \xi_v\equiv 1, \eta=0.2$                   & 0.10 & 1.47 & 1.48 & 1.44 & 0.88\\
    \hline
    F  &
    \rule{0pt}{2.5ex}
    $\ell$               & $\bc^{\tnr{FIVB}}, \xi_v\equiv 1, \eta=0.2$
    & 0.20 & 1.46 & 1.48 & 1.43 & 0.85\\
    % \hline
    % \rule{0pt}{2.5ex}
    % $\ell^{\mcL}$        & $\bc^{\tnr{FIVB}}/\sigma, \xi_v=1, \eta=0.12, s=80$   & 0.40 & 1.45 & 1.46 & 1.42 & 0.88\\
    \end{tabular}
    \caption{The metrics \eqref{U.SG}-\eqref{U.hfa-SG} obtained using the \gls{sg} algorithm using different loss functions and parameters. We always show the results with the step $\hat\mu$ obtained via \eqref{hat.mu}, except in the cases A, B and C, where we also show the results obtained using $\mu=0.01$ which is defined in the \gls{fivb} ranking.}
    \label{tab:SG.results}
\end{table}
%%%%%%%%%%%%%%%%%%%%%%%%%%%%%%%%%%%

For the algorithms based on the \gls{fivb} implicit loss function and the weighting with $\bxi^{\tnr{FIVB}}$, we evaluate the performance using the nominal adaptation step $\mu=0.01$, and, for each algorithm, we also search for the adaptation step which minimizes the validation metric overall
\begin{align}
    \label{hat.mu}
    \hat\mu & = \argmin_{\mu} \ov{U}(\mu),
\end{align}
where $\ov{U}(\mu)=\ov{U}$ shown in \eqref{U.SG}.

To indicate how much the new algorithms change the ranking when comparing to the official \gls{fivb} ranking, we calculate the average Spearman correlation coefficient
\begin{align}\label{Spearman.average}
    \ov\rho & = \frac{1}{T}\sum_{t=1}^T\rho(\btheta_t^{\tnr{FIVB}},\hat\btheta_t),
\end{align}
where $\rho(\btheta, \btheta')$ is the Spearman correlation between the skill vectors $\btheta$ and $\btheta'$.\footnote{If the order implied by the values in $\btheta$ is the same as the order implied by $\btheta'$, we have $\rho(\btheta,\btheta')=1$; if, on the other hand, $\btheta'$ is obtained by taking elements of $\btheta$ in reversed order, we have $\rho(\btheta, \btheta')=-1$.}

Note that even using exactly the same parameters as the \gls{fivb} ranking (case A in Table~\ref{tab:SG.results}), our results are not the same as the official ones because we discarded the forfeited matches; this explains why the Spearman correlation is the largest among the algorithms, but it is not perfect, \ie $\ov\rho<1$. 

Without surprise, the best result $\ov{U}=1.46$, is obtained using the true log-score $\ell_y(z)$, given by \eqref{log.score.FIVB} (case F in Table~\ref{tab:SG.results}) and, on average, this ranking is the least correlated with the \gls{fivb} ranking ($\ov\rho=0.85$); remember, however, that the implementation of \eqref{dot.log.score.FIVB} is numerically complex. The second-best result is obtained using i) the numerical scores $\tilde{r}_y(\bc^{\tnr{FIVB}})$, see \eqref{r.y.simple}, together with the constant weighting $\xi_v\equiv 1$ (case E in Table~\ref{tab:SG.results}). An improvement in performance from using the \gls{hfa} $\eta=0.2$ is slight (see cases A and B) but, quantitatively, in line with the improvement observed in home-matches in Fig.~\ref{fig:ALO_logloss} (where, after applying the \gls{hfa}, the value of $U(\bp)$ changes from $1.36$ to $1.34$).

To show an example of how the algorithms (A, E and F) affect the ranking of the top teams, we show the ranking in Table~\ref{tab:ranking_countries}, where the three front-runners stay the same, but the position of the remaining teams changes.

\begin{table}
    \centering
    \begin{tabular}{c|c|c|c|c|c|c|c}
        case & \multicolumn{7}{c}{top teams and skills}\\
        \hline
        \multirow{2}{*}{A ($\mu=0.01$)}
        & POL & USA & JPN & BRA & ITA & ARG & RUS \\
        & 423.8 & 396.8 & 345.9 & 345.0 & 344.3 & 317.0 & 315.7 \\
        \hline
        \multirow{2}{*}{A ($\mu=0.03$)}
        & POL & USA & JPN & ARG & SLO & GER & BRA \\
        & 517.4 & 478.2 & 420.2 & 374.9 & 368.1 & 361.3 & 360.2 \\
        \hline
        \multirow{2}{*}{E}
        & POL & USA & JPN & ARG & ITA & SLO & GER \\
        & 468.9 & 462.2 & 409.8 & 381.0 & 375.3 & 366.4 & 356.0 \\
        \hline
        \multirow{2}{*}{F}
        & POL & USA & JPN & GER & ARG & SLO & ITA \\
        & 528.9 & 524.3 & 475.3 & 436.9 & 430.6 & 416.3 & 401.4 
    \end{tabular}
    \caption{Ranking of the top teams in the last day of 2023 for different algorithm with parameters specified in Table~\ref{tab:SG.results}.}
    \label{tab:ranking_countries}
\end{table}

Regarding the choice of the adaptation step, we observe that 
\begin{itemize}
    \item The adaptation step $\mu=0.01$ used by the \gls{fivb} ranking is too small and, by increasing it three- or four-fold, performance improves. This can be explained using the interpretation of the \gls{sg} algorithm as a simplified Kalman filter proposed by \citet[Sec.~3.3]{Szczecinski21}, where the adaptation step in the \gls{sg} algorithm has a meaning of the posterior variance of the skills. In simple terms, \gls{fivb} is over-optimistic about the uncertainty (variance) in the estimation of the skills.
    \item Since the weighting may also be interpreted as a variable step size, by removing it, \ie using $\xi_v\equiv 1$, we have to explicitly increase the step size; this explains the large value of $\mu$ for each configuration in which we use $\xi_v\equiv 1$. 
\end{itemize}

\section{Conclusions}\label{Sec:Conclusions}

In this work, the online ranking algorithm used by the \gls{fivb} is presented in the statistical learning framework. To our best knowledge, the \gls{fivb} ranking is the first to adopt an explicit probabilistic model (here, the \acrfull{cl} model) of the multi-level ordinal outcomes, and, from the statistical perspective, this is a step in the right direction. On the other hand, the algorithms adopt simplifications that we demonstrate to be suboptimal, which is the ``misstep" in the title. However, we show how these simplifications may be easily corrected using well-defined formulas to calculate the numerical scores, see \eqref{r.y.simple}. The impact of these changes on the on-line ranking can be seen in Table~\ref{tab:SG.results} and Table~\ref{tab:ranking_countries}.

To analyze the algorithm, we use two approaches: i) the analytical, where the approximations and simplifications allow us to draw conclusions about the properties of the model, as well as to optimize its parameters, and ii) the numerical, where we explicitly optimize the parameters of the model from the outcomes of the international volleyball matches used in the \gls{fivb} ranking.

The analytical approach is easily reproducible, while the numerical optimization which relies on the cross-validation strategy allows us to validate the insights obtained analytically. This led to the following understanding of the current \gls{fivb} ranking algorithm:
\begin{itemize}
    \item The \gls{fivb} algorithm should be seen as the approximate \gls{ml} inference of the skills from the ordinal match outcomes. The approximations are due to the use of the \gls{sg} to solve the optimization problem, and, more importantly, due to the use of the loss functions, which are proxies for the log-likelihood of the \gls{ml} approach. We explain the rationale for using such proxy loss functions. 
    \item Although the form of loss functions is not explicitly mentioned in the description of the \gls{fivb} algorithm, they can be inferred from the equations, and we show how they depend on the numerical scores that are attributed to the ordinal match outcomes in the \gls{fivb} algorithm. This is interesting because in this way we explain the meaning of the numerical scores, as, from the modeling perspective, the ordinal variables do not have numerical values.
\end{itemize}

Regarding the model underlying the current \gls{fivb} ranking algorithm and the algorithm itself, we studied:
\begin{itemize}
    \item \textbf{CL model thresholds} $\bc^{\tnr{FIVB}}$, see \eqref{bc.FIVB}, which define the probabilistic ordinal model of the data. They fit relatively well the data we analyzed (the \gls{fivb} matches from 2021-2023), see Sec.~\ref{Sec:Choose.bc}.
    
    \item \textbf{Numerical scores}  $\br^{\tnr{FIVB}}$, see Table~\ref{tab:numerical.values.FIVB}, which define the algorithm. These are shown, both analytically and numerically, to be inadequately set, see Sec.~\ref{Sec:Numerical.score}.
    
    \item \textbf{Importance weights} $\bxi$, see Table~\ref{tab:FIVB.xi}, which change the contribution of the outcome of the match depending on the match type. These are shown to be irrelevant from an statistical point of view, see Sec.~\ref{Sec:categ.weighting}.
    
    \item \textbf{\Acrfull{hfa}} which deals with the matches played on the home venues by artificially boosting the skills of the home-team. We found that the \gls{hfa} is a relevant parameter that improves the prediction performance. The improvements are relatively small, which may explain why the current \gls{fivb} ranking does not use the \gls{hfa}. On the other hand, there is no cost related to its application.

    % \item \textbf{CDF function}, which defines the \gls{cl} model, see \eqref{Qy(z).FIVB}. Replacing the Gaussian \gls{cdf} used currently by the \gls{fivb} ranking with the logistic \gls{cdf}, see \eqref{Q.y.z.CL.logistic}, produces multiple benefits: i) the resulting algorithm is very simple, see \eqref{SG.CL.model} and \eqref{G.cl.y.z}, ii) it makes the concept of the numerical score superfluous, and iii) its thresholds $\bc^{\mcL}$ can be obtained though a simple transformation of the thresholds $\bc^{\tnr{FIVB}}$ currently used in the \gls{fivb} ranking, see \eqref{Phi.approx.mcL.z.eta.c}.

\end{itemize}

\textbf{Recommendations}

In summary, by keeping the structure of the current \gls{fivb} ranking and by exploring the optimality of the above model/algorithm choices, we came up with new parameters that can improve the performance of the algorithm. And, since \gls{fivb} explicitly says that its algorithm may be updated, if this is to happen, our recommendations, in order of importance, are the following:
\begin{enumerate}
    \item \label{recommendation.point.scores} Change the numerical score $\br^{\tnr{FIVB}}$ to be similar to those suggested in \eqref{tilde.r.for.FIVB}-\eqref{tilde.r.for.FIVB.2}.

    \item Introduce the \gls{hfa} to the algorithm. This is a simple and low-cost modification, yielding an improvement in the prediction of the matches played on the home-venue.
    
    \item Remove the weighting of the matches with $\xi^{\tnr{FIVB}}_v$. Or, if its use is motivated by some extra-statistical (\eg entertainment) reasons, decrease the differences between the possible values of $\xi_v^{\tnr{FIVB}}$.

    % \item Use the logistic \gls{cdf} in the \gls{cl} model, as this removes the need for auxiliary numerical scores (thus making the recommendation \#\ref{recommendation.point.scores} void) and yields a simple, and easily interpretable algorithm.
\end{enumerate}

\textbf{Further work}

While in this work, we focus on the \gls{fivb} ranking, the evaluation methodology we propose can be used more broadly, to analyze the ranking algorithms. In particular, the ``reverse engineering'' approach we used to reveal the form of the (implicit) objective loss function (see Sect.~\ref{Sec:Implicit.Loss.Function}) is particularly useful. It can be applied to analyze the sub-optimality of the ranking algorithms which, in practice, may be defined without an explicit probabilistic model. In that sense, our evaluation methodology is more general than reverse engineering, which was used in \cite{Szczecinski22a} to unveil the model underlying the \gls{fifa} ranking.

Beyond this general recommendation and focusing specifically on improving the \gls{fivb} ranking, the following venues can be explored:
\begin{itemize}
    \item Considering a time-variant model for the skills in the design of the algorithm, \eg using ideas already shown before in \citep{Fahrmeir92}, \citep{Glickman93_thesis}, \citep{Knorr00}, \citep{Szczecinski21}. It may require particular attention, as the current version of the \gls{fivb} ranking algorithm is not a straightforward implementation of the \gls{ml} ranking.
    \item Analyzing alternative ordinal models \cite{McCullagh80}, taking into account the simplicity of the algorithm they produce, including the use of different \gls{cdf} in the model \eqref{Qy(z).FIVB} \cite[Ch.~9.1.3]{Tutz12_book}\cite[Ch.~8.3]{Agresti13_book}, or the \gls{ac} models \citep[Ch.~9.4.5]{Tutz12_book} \cite{Szczecinski22}.
\end{itemize}

% \appendix
\begin{appendices}
%%%%%%%%%%%%%%%%%%%%%%%%%%%%%%%%%%%
\section{Notation in FIVB ranking}\label{Sec:FIVB.notation}

We show in Table~\ref{tab:Equivalence}, the relationship between our notation and the one used in the \gls{fivb} ranking description \citep{fivb_rating}, 
where the following abbreviations are used
\begin{itemize}
    \item WR: World ranking (here, a \gls{fivb} ranking)
    \item WRS: World ranking score (here, we call it skills $\btheta_{m,t}$)
    \item SSV: Set score variant (we call it numerical score $r_y$)
    \item EMR: Expected match result (here, the expected score $\check{r}(z)$, see \eqref{check.vt}) 
    \item MWF: Match weighting factor (here, it corresponds to $10\xi_{v_t}$) 
    \item Scaled difference between WRSs $\Delta=8(\tnr{WRS1}-\tnr{WRS2})/1000$
\end{itemize}

\begin{table}[ht]
    \centering
    \begin{tabular}{c|c}
        Our notation  & \gls{fivb} notation\\
        \hline
        \hline
        $\theta_{m,t}, \theta_{n,t}$ & WRS1, WRS2\\
        \hline
        $c_0^{\tnr{FIVB}},\ld,c_4^{\tnr{FIVB}}$ & C1, \ldots, C5\\
        \hline
        $z_t/s = (\theta_{m,t} - \theta_{n,t})/s$ & $\Delta=8(\tnr{WRS1}-\tnr{WRS2})/1000$\\
        \hline
        $s$     &  1000/8=125\\
        \hline
        $\mfP_0(z_t),\ld,\mfP_5(z_t) $ & P1, \ldots, P5 \\
        \hline 
        $\Phi(z)$ & $\sim N(0,1)(z)$\\
        \hline
        $r_{y_t}^{\tnr{FIVB}}$  &  SSV\\
        \hline
        $\check{r}(z_t/s)$  &   EMR\\
        \hline
        $10\xi_{v_t}$   &   MWF\\
        \hline
        $-g^{\tnr{FIVB}}(z_t/s)=r_{y_t}^{\tnr{FIVB}} - \check{r}(z_t/s)$   & WR value \\
        \hline
        $\theta_{m,t+1}-\theta_{m,t}=-10\xi_{v_t}g^{\tnr{FIVB}}(z_t/s)$  & WR points = WR values * MWF /8 
    \end{tabular}
    \caption{Equivalence of this work's notation and the one used in the description of the \gls{fivb} ranking.}
    \label{tab:Equivalence}
\end{table}

With this notation, the update formula is given by
\begin{align}
    \text{WRS1} \leftarrow  \text{WRS1} + \text{WR points}
\end{align}
and corresponds to \eqref{FIVB.algorithm} which, focusing on the update of the skills of the home team $m$, may be written as
\begin{align}
\label{FIVB.algorithm.m}
    \theta_{m,t+1} & 
    =
    \theta_{m,t} - \mu s \xi_{v_t}g^{\tnr{FIVB}}_{y_t}(z_t/s).
\end{align}

%%%%%%%%%%%%%%%%%%%%%%%%%%%%%%%%%%%
\section{Optimization of the cross-validation metric}\label{Sec:estimation.hyper}

The simplest optimization of the cross-validation metric $U(\bp)$ may be done via the steepest descent
\begin{align}
    \hat\bp\leftarrow \hat\bp - \kappa\nabla_{\bp} U(\bp),
\end{align}
where, $\kappa$ is the step-size, and to calculate the gradient $\nabla_{\bp} U(\bp)$, we have to calculate derivatives of $U(\bp)$ with respect to a parameter $q\in\bp$. This can be done as follows:
\begin{align}
\label{dU.dq}
    \frac{\partial}{\partial q}U(\bp)
    &=
    \frac{1}{T}\sum_{t=1}^T \frac{\partial}{\partial q}\ell^{\tnr{val}}_{y_t}( \hat{z}_{t,\backslash{t}},\bp)\\
    \frac{\partial}{\partial q}\ell^{\tnr{val}}_{y_t}( \hat{z}_{t,\backslash{t}})
    &=
    \frac{\partial \hat{z}_{t,\backslash{t}}}{\partial q}\dot\ell^{\tnr{val}}( \hat{z}_{t,\backslash{t}},\bp)
    +
    \frac{\partial}{\partial q}\ell^{\tnr{val}}_{y_t}( \hat{z}_{t,\backslash{t}},\bp)\\
\label{dz.t.t.dq}
    \frac{\partial \hat{z}_{t,\backslash{t}}}{\partial q}
    &=
    \frac{\partial \hat{z}_{t}}{\partial q}
    +
    \frac{\partial }{\partial q}
    \left[
    \frac{\xi_{v_t}\dot\ell_{y_t}(\hat{z}_t,\bp) a_t}{1-\xi_{v_t}\ddot\ell_{y_t}(\hat{z}_t,\bp)a_t}
    \right].
\end{align}

In \eqref{dz.t.t.dq}, we will need
\begin{align}
\label{dzt.dq}
    \frac{\partial \hat{z}_t}{\partial q}
    &=
    \bx\T_t\frac{\partial \hat{\btheta}}{\partial q}\\
\label{dat.dq}
    \frac{\partial a_t}{\partial q}
    &=
    \bx\T_t\frac{\partial \hat{\matH}^{-1}}{\partial q}\bx_t = -\bx\T_t\hat{\matH}^{-1}\frac{\partial \hat{\matH}}{\partial q}\hat{\matH}^{-1}\bx_t\\
\label{dH.dq}
    \frac{\partial \hat\matH}{\partial q}
    &=
    \sum_{t=1}^T\bx_t \frac{\partial}{\partial q}\big[\xi_{v_t}\ddot\ell_{y_t}(\hat{z}_t,\bp)\big]\bx\T_t +\IND{q=\gamma}\matI\\
\label{d.ell.dq}
    \frac{\partial}{\partial q}\dot\ell_{y_t}(\hat{z}_t,\bp)
    &= 
    \frac{\partial \hat{z}_t}{\partial q}\ddot\ell_{y_t}(\hat{z}_t,\bp) +  
    \frac{\partial}{\partial q}\dot\ell_{y_t}(\hat{z}_t,\bp)
    \\
\label{d2.ell.dq}
    \frac{\partial}{\partial q}\ddot\ell_{y_t}(\hat{z}_t,\bp)
    &= \frac{\partial \hat{z}_t}{\partial q}\dddot{\ell}_{y_t}(\hat{z}_t,\bp) +  
    \frac{\partial }{\partial q}\ddot\ell_{y_t}(\hat{z}_t,\bp)
\end{align}
where, in \eqref{dat.dq} we used \citep[Eq.~(40)]{Petersen12_book}, and $\dddot\ell_{y}(z,\bp)=\frac{\partial^3}{\partial z^3} \ell_{y}(z,\bp)$.

From implicit function theorem, see \citep[Th.~1]{Lorraine19}, using $\hat{\btheta}\equiv\hat{\btheta}(\bp)$
\begin{align}
    \bzero &=
    \frac{\partial}{\partial q}\Big[\nabla_{\btheta}J(\hat\btheta(\bp),\bp)\Big]\\
    \bzero&=\nabla^2_{\btheta}J(\hat\btheta(\bp),\bp) \frac{\partial \hat\btheta(p)}{\partial q} + \frac{\partial}{\partial q}\nabla_{\btheta}J(\hat\btheta,\bp)\\
\label{d.theta.d.q.implicit}
    \frac{\partial \hat{\btheta}}{\partial q}
    &= -\hat{\matH}^{-1} \frac{\partial}{\partial q} \nabla_{\btheta}J(\hat\btheta,\bp)\\
    \frac{\partial}{\partial q}\nabla_{\btheta}J(\hat\btheta,\bp)
    &=
    \sum_{t=1}^T \frac{\partial}{\partial q} \big[\xi_{v_t}\dot\ell_{y_t}(\bx_t\T\hat{\btheta}, \bp)\big]\bx_t + \IND{\gamma=q}\hat\btheta.
\end{align}

By plugging $\hat{z}_{t,\backslash{t}}(\bp)$ into \eqref{U(p).define} we obtain the function $U(\bp)$ which depends on $\bp$ and we can calculate the gradient $\nabla_{\bp}U(\bp)$.

Similarly, we can use the Newton method
\begin{align}
    \hat\bp\leftarrow \hat\bp - [\nabla^2_{\bp}U(\bp)]^{-1}\nabla_{\bp}U(\bp)
\end{align}
where, to calculate the Hessian, $\nabla^2_{\bp}U(\bp)$ we need second order derivatives.

However, the expressions for the gradient and, especially, the Hessian, quickly become cumbersome; see \citep{Burn20}. Thus, instead of explicit differentiation, we use the automatic differentiation available in JAX and JAXopt python-compliant packages \citep{jax2018github}, \citep{Blondel21} with a particularly interesting feature which automatically finds the implicit differentiation required to find the derivative of $\hat\btheta$ with respect to hyperparameters in $\bp$ as specified by \eqref{d.theta.d.q.implicit}.

We do not show more details to not overcomplicate the presentation, especially that they are not really required because the numerical optimization is used to confirm the observation we made using the analytical insight. In fact, the performance of the online algorithm shown in Sec.~\ref{Sec:Real.time.ranking} uses the parameters (shown in Table~\ref{tab:SG.results}) which are explicitly defined prior to the application of the algorithms.
\end{appendices}

\ifdefined\ARXIV
\input{\CFilesBib/output.bbl.my}
\else
\bibliographystyle{\CFilesBib/DeGruyter}
\bibliography{\CFilesBib/IEEEabrv,\CFilesBib/references_all,\CFilesBib/references_rank}
\fi

\end{document}